%% file: main.tex
\documentclass[acmtog, nonacm]{acmart}
\acmSubmissionID{296}

\usepackage{booktabs} 

\input{macros.tex}
\usepackage{graphicx}
\usepackage{booktabs}
\usepackage{nicefrac}
\usepackage{xfrac}
\usepackage{calc}
\usepackage{soul}
\usepackage{mathtools}
\usepackage[percent]{overpic}
\usepackage{adjustbox}
\usepackage{colortbl}
\usepackage{cancel}
\usepackage{float}

\newcommand{\MYhref}[3][blue]{\href{#2}{\color{#1}{#3}}}%

\usepackage[ruled]{algorithm2e} 

\SetAlFnt{\small}
\SetAlCapFnt{\small}
\SetAlCapNameFnt{\small}
\SetAlCapHSkip{0pt}

\acmJournal{TOG}

\begin{document}
\title{NeRF-Casting: Improved View-Dependent Appearance with Consistent Reflections}

\author{Dor Verbin}
\affiliation{%
  \institution{Google} \country{USA}}
\author{Pratul P. Srinivasan}
\affiliation{%
  \institution{Google} \country{USA}}
\author{Peter Hedman}
\affiliation{%
  \institution{Google} \country{UK}}
\author{Ben Mildenhall}
\affiliation{%
  \institution{Google} \country{USA}}
\author{Benjamin Attal}
\affiliation{%
  \institution{Carnegie Mellon University} \country{USA}}
\author{Richard Szeliski}
\affiliation{%
  \institution{Google} \country{USA}}
\author{Jonathan T. Barron}
\affiliation{%
  \institution{Google} \country{USA}}

\renewcommand\shortauthors{Verbin et al.}

\authorsaddresses{}

\begin{abstract}

Neural Radiance Fields (NeRFs) typically struggle to reconstruct and render highly specular objects, whose appearance varies quickly with changes in viewpoint. Recent works have improved NeRF's ability to render detailed specular appearance of distant environment illumination, but are unable to synthesize consistent reflections of closer content. Moreover, these techniques rely on large computationally-expensive neural networks to model outgoing radiance, which severely limits optimization and rendering speed. We address these issues with an approach based on ray tracing: instead of querying an expensive neural network for the outgoing view-dependent radiance at points along each camera ray, our model casts reflection rays from these points and traces them through the NeRF representation to render feature vectors which are decoded into color using a small inexpensive network. We demonstrate that our model outperforms prior methods for view synthesis of scenes containing shiny objects, and that it is the only existing NeRF method that can synthesize photorealistic specular appearance and reflections in real-world scenes, while requiring comparable optimization time to current state-of-the-art view synthesis models.
\end{abstract}

%
%


%
%


\newcommand{\teaserwidth}{0.245\linewidth}

\begin{teaserfigure}
\begin{center}
  {\Large Interactive webpage at \MYhref{https://nerf-casting.github.io}{\texttt{https://nerf-casting.github.io}}}\\
    \begin{tabular}{@{}c@{\,\,}c@{\,\,}c@{\,\,}c@{}}
    \adjincludegraphics[Clip={{0.548\width} {0.473\height} {0.224\width} {0.243\height}},width=\teaserwidth]{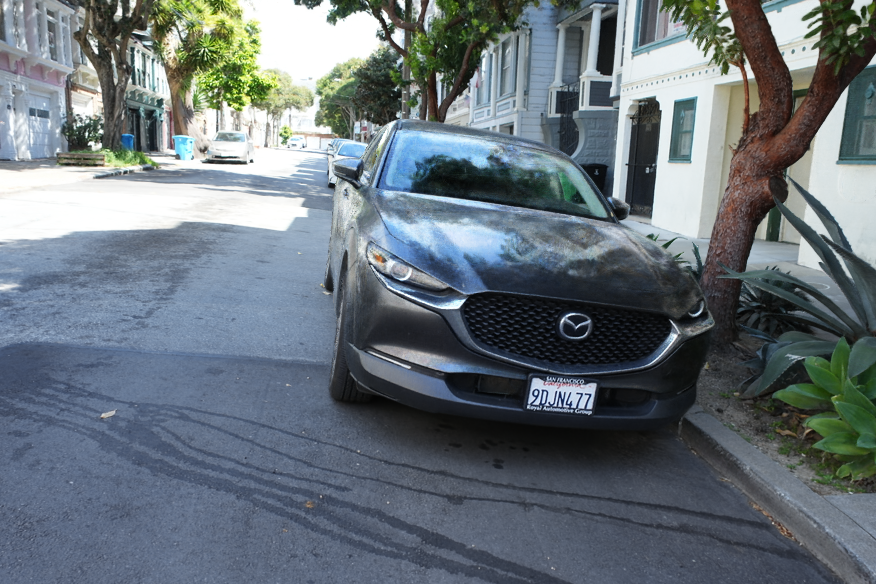} &
    \adjincludegraphics[Clip={{0.548\width} {0.473\height} {0.224\width} {0.243\height}},width=\teaserwidth]{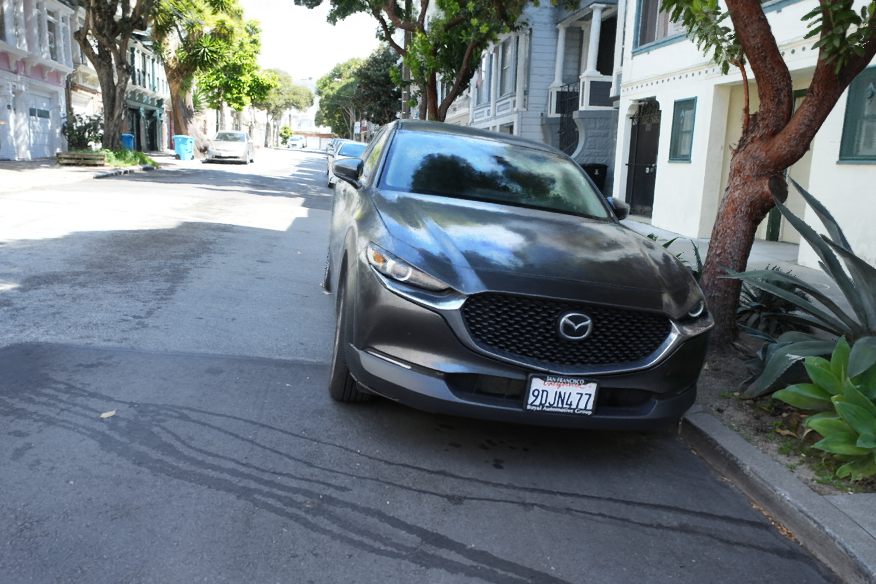} &
    \adjincludegraphics[Clip={{0.548\width} {0.473\height} {0.224\width} {0.243\height}},width=\teaserwidth]{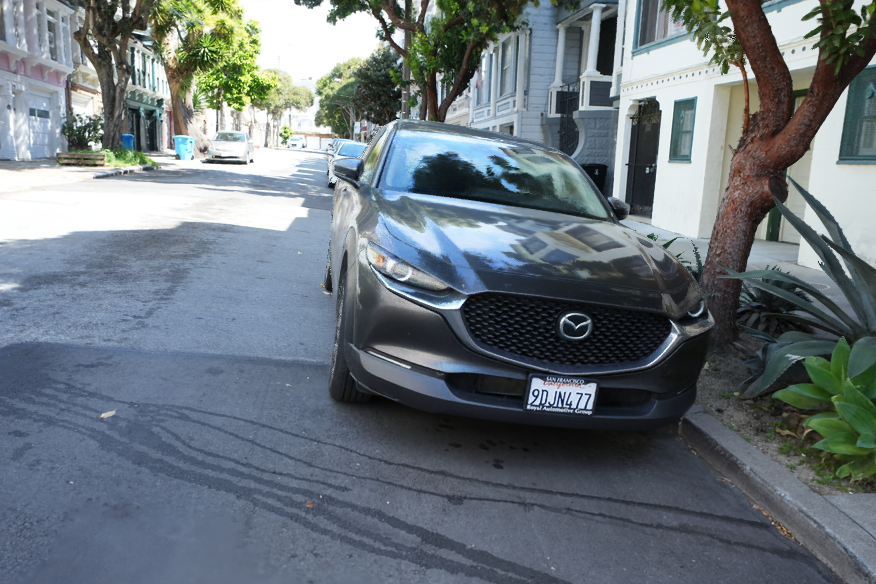} &
    \adjincludegraphics[Clip={{0.548\width} {0.473\height} {0.224\width} {0.243\height}},width=\teaserwidth]{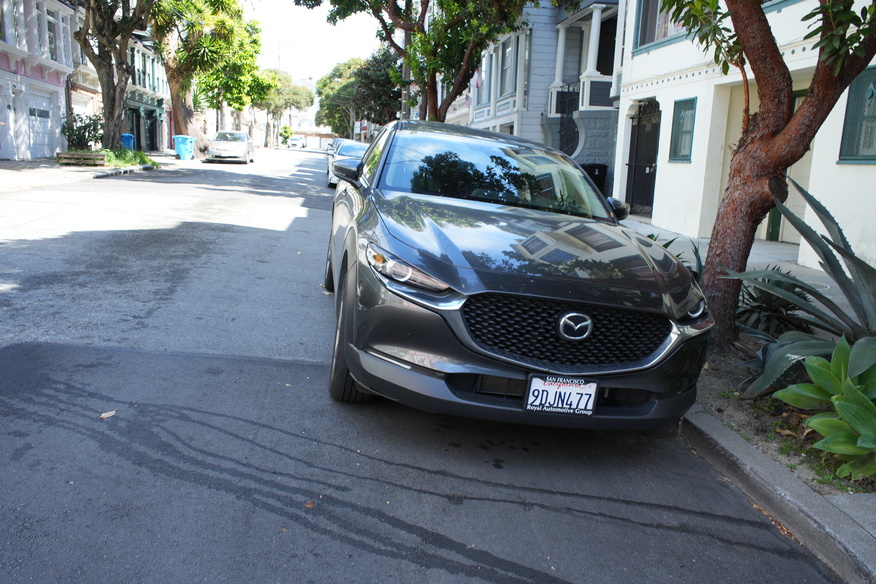}\\
    \adjincludegraphics[width=\teaserwidth]{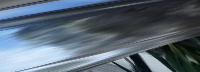} &
    \adjincludegraphics[width=\teaserwidth]{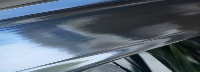} &
    \adjincludegraphics[width=\teaserwidth]{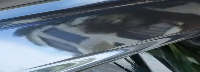} &
    \adjincludegraphics[width=\teaserwidth]{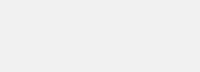}\\
    \small (a) Zip-NeRF~\cite{barron2023zipnerf} & \small (b) UniSDF~\cite{wang2023unisdf} & \small (c) Ours & \small (d) Ground truth
    \end{tabular}
    \caption{Our method, NeRF-Casting, recovers and renders scenes with higher-quality reflections than prior work. Here we show cropped renderings and a ground-truth image for a scene in our dataset (top), as well as a slice from a video rendering along a smooth camera path (one horizontal row of pixels stacked vertically across multiple video frames) to emphasize the continuity of high-frequency reflected content in our result (bottom). Please see the videos provided in the supplemental webpage to observe the accurate motion of our rendered reflections.}
  \label{fig:teaser}
\end{center}
\end{teaserfigure}

\maketitle

\input{content}

\end{document}

%% file: macros.tex
\usepackage{xspace}

\makeatletter
\DeclareRobustCommand\onedot{\futurelet\@let@token\@onedot}
\def\@onedot{\ifx\@let@token.\else.\null\fi\xspace}

\def\ie{\emph{i.e}\onedot}

\def\etal{\emph{et al}\onedot}
\makeatother

\definecolor{yellow}{rgb}{1, 1, 0.7}
\definecolor{orange}{rgb}{1, 0.85, 0.7}
\definecolor{red}{rgb}{1, 0.7, 0.7}
\definecolor{darkgreen}{rgb}{0.0, 0.7, 0.0}
\definecolor{todored}{rgb}{1, 0.2, 0.2}

\newcommand{\vb}{\mathbf{b}}
\newcommand{\vf}{\mathbf{f}}
\newcommand{\vo}{\mathbf{o}}
\newcommand{\vr}{\mathbf{r}}
\newcommand{\vt}{\mathbf{t}}
\newcommand{\vx}{\mathbf{x}}

\newcommand{\vomega}{\boldsymbol{\omega}}

\newcommand{\avgx}{\bar{\vx}}
\newcommand{\avgf}{\bar{\vf}}
\newcommand{\avgnormal}{\bar{\normal}}
\newcommand{\avgkappa}{\bar{\kappa}}

\newcommand{\rgb}{\mathbf{c}}
\newcommand{\normal}{\mathbf{n}}
\newcommand{\viewdir}{\mathbf{d}}
\newcommand{\refdir}{\mathbf{d}'}
\newcommand{\density}{\tau}
\newcommand{\roughness}{\rho}
\newcommand{\scalemult}{\gamma}

\newcommand\lft{\mathopen{}\left}
\newcommand\rgt{\aftergroup\mathclose\aftergroup{\aftergroup}\right}

\newcommand{\jacobian}{\mathbf{J}}

\newcommand{\stopgrad}{\operatorname{sg}}

%% file: content.tex
\section{Introduction}

Neural Radiance Fields (NeRFs) represent a 3D scene's underlying geometry and color using neural networks, for the goal of rendering novel unobserved views of the scene. Early variants of NeRF used multi-layer perceptrons (MLPs) to map from 3D coordinates to volume densities and view-dependent colors. However, the large MLPs needed to represent detailed 3D geometry and color are extremely slow to train and evaluate. Recent work has focused on making NeRF more efficient by replacing large MLPs with voxel-grid-like data structures or combinations of grids and small MLPs. These methods have enabled scaling NeRF to representing detailed large-scale scenes, but their benefits are limited to 3D geometry and mostly \emph{diffuse} color.

Scaling NeRF's ability to model realistic view-dependent appearance is still a challenge. State-of-the-art models for view synthesis of shiny objects with high-frequency view-dependent appearance are limited in two ways. First, they are only able to synthesize accurate reflections of distant environment illumination and struggle to render convincing reflections of nearby scene content. Second, they rely on large MLPs to represent the view-dependent outgoing radiance at any point and struggle to scale to larger real-world scenes with detailed reflections.

We present an approach that solves both of these issues by introducing ray tracing into NeRF's rendering model. Instead of querying an expensive MLP for the view-dependent appearance at each point along a camera ray, our method casts reflection rays from these points into the NeRF geometry, samples properly anti-aliased features from the reflected scene content, and uses a small MLP to decode these features into reflected color. Casting rays into the recovered NeRF naturally synthesizes consistent reflections of nearby and distant content. Moreover, computing appearance by ray tracing reduces the burden of representing highly-detailed view-dependent functions at each point in the scene with a large MLP.

Our approach is both more efficient than prior work for improving NeRF's view-dependent appearance and renders higher quality specular reflections, particularly of nearby content.

\section{Related Work}

A Neural Radiance Field~\cite{mildenhall2020nerf} consists of an MLP that maps from a 3D coordinate within the scene to the corresponding volume density and view-dependent color at that position. NeRF renders rays passing through a scene by querying the MLP at sampled points along each ray and integrating the returned volume densities and colors into a single color. Evaluating an MLP hundreds of millions of times to render a single image can be prohibitively slow, so recent research has focused on making NeRF more efficient and scalable. In particular, many effective methods trade compute for storage and leverage data structures based on voxel grids to accelerate NeRF's optimization and rendering. These include standard voxel grids~\cite{relufields, dvgo, fridovich2022plenoxels}, low-rank tensor factorizations~\cite{tensorf}, and voxel pyramids with hash maps~\cite{mueller2022instant}. These methods are primarily focused on scaling NeRF's representation of 3D volume density to represent larger and more detailed scenes. However, they do not usually modify NeRF's representation of view-dependent color, which is a 5D function of both position and viewing direction.

A separate line of work has focused on improving NeRF's ability to reconstruct and render highly specular (or shiny) content. Ref-NeRF~\cite{verbin2022refnerf} demonstrated that reparameterizing outgoing radiance as a function of the reflected view direction is effective in cases where geometry is estimated accurately. However, Ref-NeRF's parameterization of outgoing radiance is most effective for objects that are mainly illuminated by distant light sources, which is often the case for synthetically-generated scenes. Furthermore, the outgoing radiance at every point must be encoded in the weights of an MLP. In the case of highly glossy objects, this MLP must have a large enough capacity to represent the entire environment viewed from any point in the scene, and is therefore costly to evaluate. Other works that focus on recovering specular appearance with NeRF also adopt Ref-NeRF's reparameterization of view-dependent appearance. ENVIDR~\cite{liang2023envidr} pre-trains the MLP that maps from reflection direction to color to effectively encode a prior on view-dependent appearance. UniSDF~\cite{wang2023unisdf} improves Ref-NeRF's geometry representation to obtain smoother normals that are better for rendering surface reflection. SpecNeRF~\cite{specnerf} modifies Ref-NeRF's encoding of view directions to vary spatially according to a set of optimized 3D Gaussians, which helps the view-dependent color network model reflections of nearby content. However, this encoding does not ensure accurate reflections when light sources are occluded. NeAI~\cite{zhuang2023neai} traces reflection cones within an MLP-based NeRF to compute specular appearance, but requires ground-truth geometry and masks that separate the object from the environment in order to supervise the MLP that represents reflected lighting. Work concurrent to ours by Wu \etal~\shortcite{liwen} proposes a Neural Directional Encoding that also traces cones through the NeRF model to simulate near-field reflections. However, they rely on a large MLP to represent geometry, which is slow to train and render compared to more modern grid-based representations. Furthermore, they primarily focus on synthetic scenes with clearly separated near-field and distant content.

The methods discussed above address the same problem as our work: improving view-dependent appearance in view synthesis. An alternative approach to synthesizing specular appearance is inverse rendering: estimating representations of scene materials and lighting, alongside geometry. Many existing works combine NeRF with inverse rendering and explicitly model light transport through the 3D scene to recover representations that can render specular appearance~\cite{bi2020nrf,jin2023tensoir,srinivasan2021nerv,mai2023neural}. However, these techniques must perform expensive Monte Carlo integration over the entire hemisphere of incoming lighting (using hundreds to thousands of sampled reflection rays) to estimate outgoing radiance at any point in the scene, and they are typically orders of magnitude slower than standard NeRF models. Our proposed method specifically focuses on rendering specular appearance for view synthesis and is able to render accurate reflections with low computational overhead compared to existing view synthesis models.

\section{Preliminaries and Notation} \label{sec:prelims}

Neural Radiance Fields~\cite{mildenhall2020nerf} uses volume rendering to combine the densities and emitted view-dependent radiance values at samples along rays:
\begin{align} \label{eq:nerfcolor}
    \bar{\mathbf{c}} &= \sum_{i} w^{(i)} \rgb^{(i)} \,,\\ 
    w^{(i)}\!&=\!\lft(1 - \exp\lft(-\density^{(i)}\lft(t^{(i+1)} - t^{(i)}\rgt)\rgt) \rgt)\!\exp\lft(-\!\sum_{j < i} \density^{(j)} \lft(t^{(j+1)} - t^{(j)}\rgt)\rgt) \nonumber
\end{align}
where $t^{(i)}$ is the $i$th sample along the ray, and $\density^{(i)}$ and $\mathbf{c}^{(i)}$ are the volume density and color (respectively) at the $i$th sample location, $\vo+t^{(i)}\viewdir$, where $\vo$ is the ray's origin and $\viewdir$ its direction. 
Throughout this paper we denote the volume-rendered values with a ``bar'', similar to $\bar{\mathbf{c}}$ on the left of Equation~\ref{eq:nerfcolor}.

Mip-NeRF~\cite{barron2021mipnerf} replaces infinitesimal rays with pixel cones with origin $\vo$, direction $\viewdir$ and radius $\dot{r}$ (determined by the pixel footprint on the image plane). Casting cones instead of rays prevents aliasing in the scene representation and as a result, prevents aliasing in the rendered images. To represent large-scale unbounded scenes, mip-NeRF 360~\cite{barron2022mipnerf360} allocates higher representational capacity to points near the cameras using a spatial contraction function that maps all points in $\mathbb{R}^3$ to a sphere of radius $2$:
\begin{equation} \label{eq:contraction}
    \mathcal{C}(\vx) = \begin{cases}\vx \quad &\text{if } \|\vx\| \leq 1 \,, \\ \lft(2 - \frac{1}{\|\vx\|}\rgt)\frac{\vx}{\|\vx\|} \quad &\text{if } \|\vx\| > 1 \,. \end{cases}
\end{equation}

In order to represent density and color, Zip-NeRF accelerates mip-NeRF 360 by replacing its large MLP with a hash encoding, based on Instant Neural Graphics Primitives~\cite{mueller2022instant}, followed by a smaller MLP. To prevent aliasing, Zip-NeRF replaces point samples of the hash encoding grid with their expectation over 3D Gaussians. Since these expected values cannot be computed exactly, they are approximated using two components: multisampling and downweighting. Multisampling replaces each Gaussian with multiple smaller Gaussians, and downweighting multiplies each of the gathered features at these samples by a multiplier designed to prevent voxels much smaller than the Gaussian from affecting the features. The contraction function in Equation~\ref{eq:contraction} must be taken into account when computing the appropriate amount of downweighting: the determinant of the Jacobian of $\mathcal{C}$, $\det J_\mathcal{C}$, ``deflates'' the Gaussians' volumes, so Zip-NeRF downweights the grid resolution $n_\ell$ by $\operatorname{erf}\lft(\lft(\sqrt{8}\sigma(\vx) n_\ell \rgt)^{-1}\rgt)$, where:
\begin{equation} \label{eq:zipnerfscale}
    \sigma(\vx) = \gamma\cdot \dot{r}\|\vx-\vo\| \sqrt[3]{\det J_\mathcal{C}(\vx)} \,,
\end{equation}
where $\gamma$ is a fixed hyperparameter.

\section{Model}

We design our method with three goals in mind: First, we want to model accurate detailed reflections without relying on computationally expensive MLP evaluations. Second, we would like to only cast a small number of reflected rays. Third, we would like to minimize the computation required to query our representation at each point along these reflection rays.

\begin{figure}[t]
\centering
\includegraphics[width=1\linewidth]{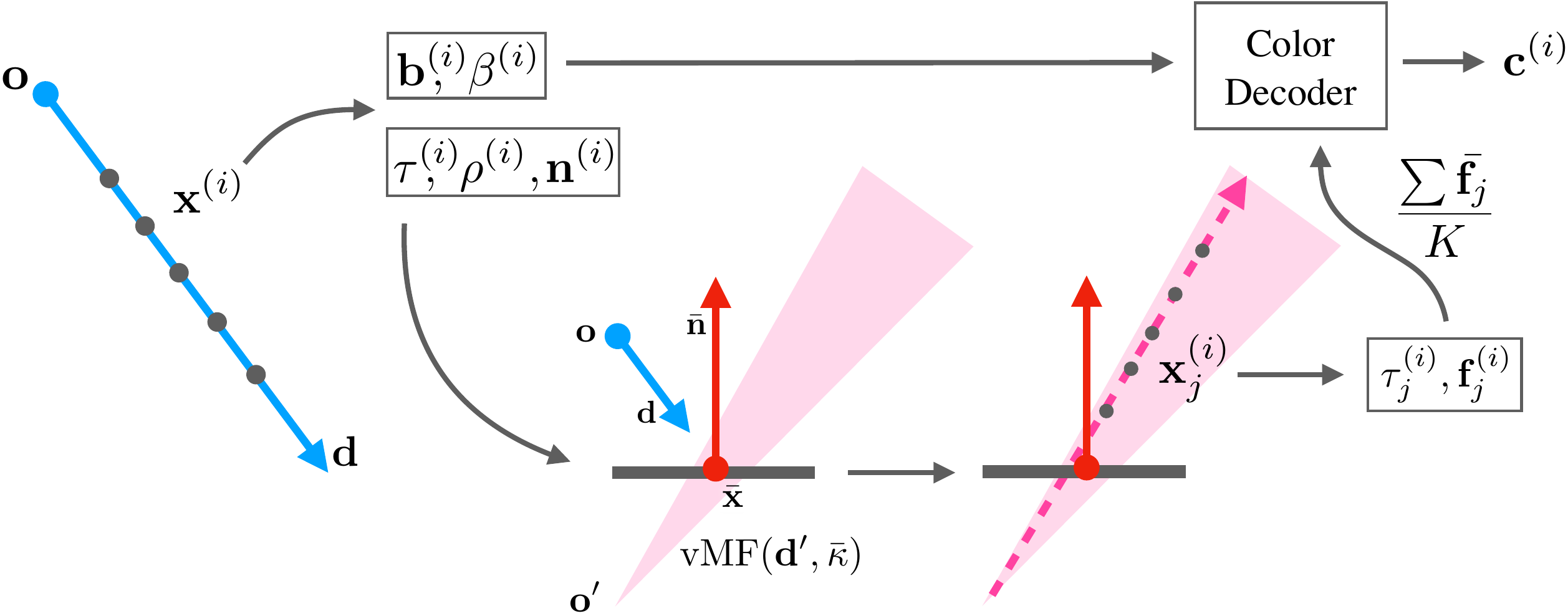}
\caption{Our model architecture for rendering a single ray with origin $\vo$ and direction $\viewdir$.
We sample $N$ points $\mathbf{x}^{(i)}$ along the ray and use a spatial encoder based on Zip-NeRF to encode each point into density $\tau^{(i)}$, roughness $\rho^{(i)}$, and surface normal $\mathbf{n}^{(i)}$. These are alpha composited to compute a single expected termination point $\bar{\mathbf{x}}$, a von Mises-Fisher distribution (vMF) width $\bar{\kappa}$, and surface normal $\bar{\mathbf{n}}$. Then $\viewdir$ is reflected around that surface to construct a vMF distribution over reflected rays $\operatorname{vMF}(\mathbf{d}', \bar{\kappa})$. We sample $K$ reflected rays ($K=5$) with location $\mathbf{o}'$ and directions $\mathbf{d}'_j$ (as in Figure~\ref{fig:refrays}). These $K$ rays are then cast, and points along them are encoded with the same model as the initial ray into $N'$ densities $\tau_j^{(i)}$ and features $\mathbf{f}_j^{(i)}$. These features are alpha composited along each ray to get per-ray features $\bar{\mathbf{f}}_j$, and the composited features are averaged into a single reflection feature $\vf$. This feature is broadcast over the original ray's samples and passed, along with bottleneck features $\mathbf{b}^{(i)}$, mixing coefficients $\beta^{(i)}$, and viewing direction $\mathbf{d}$, to the color decoder to produce RGB colors $\mathbf{c}^{(i)}$ for each point along the ray. These colors are alpha composited to render a pixel color $\bar{\mathbf{c}}$.}
\label{fig:pipeline}
\end{figure}

Our representation of 3D volume density and features is based on Zip-NeRF. As described in Section~\ref{sec:prelims}, Zip-NeRF uses a multi-scale hashgrid to store 3D features, a small MLP (1 layer, width 64) to decode these features into density, and a larger MLP (3 layers, width 256) to decode these features into color. This means that querying the density and feature for samples along rays is cheap relative to evaluating color. Considering these constraints, we propose the following procedure to render specular appearance:
\begin{enumerate}
    \item Query volume density along each camera ray to compute the ray's expected termination point and surface normal.
    \item Cast a reflected cone through the expected termination point in the reflection direction.
    \item Use a small MLP to combine the accumulated reflection feature with other sampled quantities (such as the diffuse color features and per-sample blending weights) to produce a color value for each sample along the ray.
    \item Alpha composite these samples and densities into the final color.
\end{enumerate}
These steps are illustrated in Figure~\ref{fig:pipeline} and explained in more detail below.

Section~\ref{sec:cone} describes how our method estimates the cone of reflection rays to be traced. Section~\ref{sec:generalenc} details how our model traces a small set of rays within this cone and computes volume density and features at sampled points along these rays. Section~\ref{sec:appearance} presents the complete appearance model that decodes the accumulated reflection feature into a specular color. Finally, Section~\ref{sec:geometry} provides additional details describing how our model of 3D geometry and features differs from the standard Zip-NeRF architecture.

\begin{figure}[t]
\centering
\includegraphics[width=1\linewidth, trim={0mm 7.2in 6.2in 0mm}, clip]{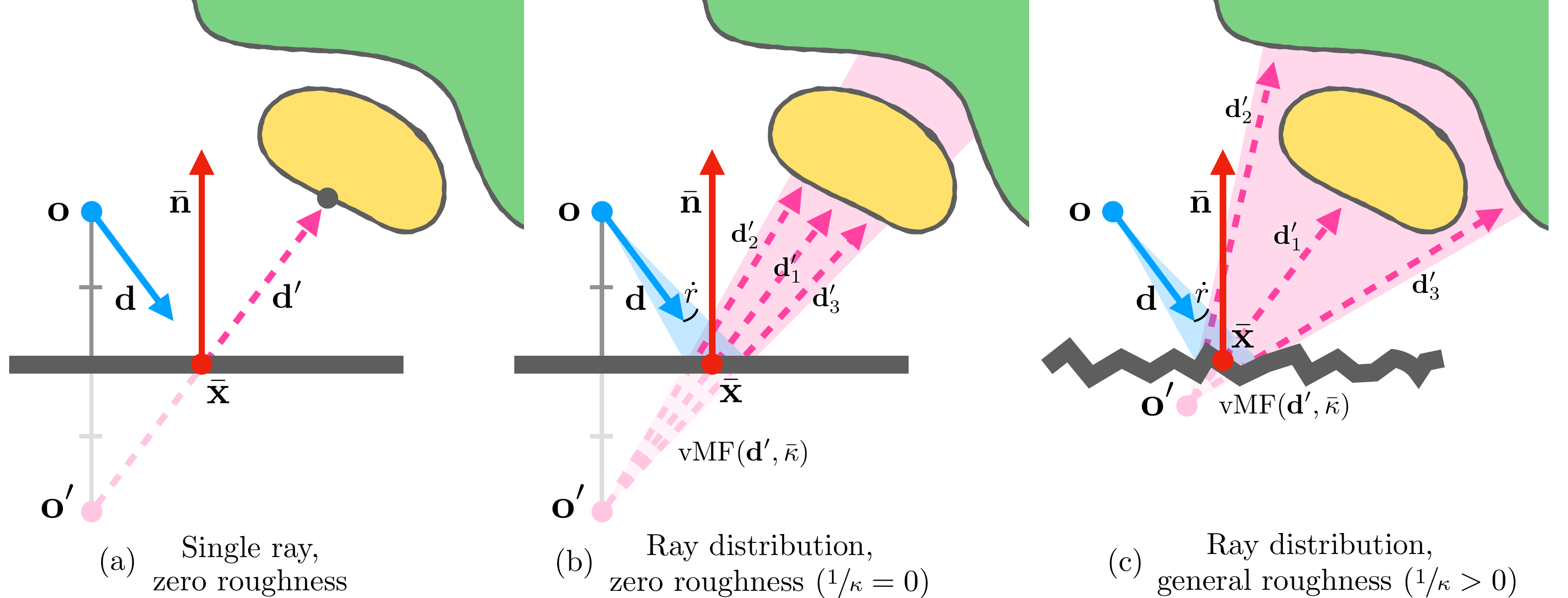}
\caption{
We visualize the reflection cone tracing procedure described in Section~\ref{sec:cone}. (a) A basic model for reflections: a ray cast from a camera ray origin $\mathbf{o}$ along view direction $\mathbf{d}$ that terminates at a coordinate $\bar{\mathbf{x}}$ with a surface normal $\bar{\mathbf{n}}$ is ``mirrored'' around the surface tangent to yield a reflected ``camera ray origin''  $\mathbf{o}'$ and direction $\mathbf{d}'$. This model does not consider that Zip-NeRF traces pixel cones instead of camera rays, which we address by (b) casting a reflection \emph{cone} with radius $\dot r$ instead of a ray, and parameterizing not just a single reflected ray direction but a von Mises-Fisher distribution over reflected rays $\operatorname{vMF}(\mathbf{d}', \bar{\kappa})$, where $\bar{\kappa}$ is the inverse-roughness of the surface. We approximate this distribution with a small set of rays $\{ \mathbf{d}'_j \}$. (c) When the surface is not a perfect mirror, the reflection cone widens and the origin of the reflected ray cone must be pulled closer to the surface location to maintain the same intersection with the pixel cone.
}
\label{fig:refrays}
\end{figure}

\subsection{Reflection Cone Tracing} \label{sec:cone}

For a given ray with origin $\vo$ and direction $\viewdir$, we sample $N$ points $\{\vx^{(i)}\}_{i=1}^N$ along it, and evaluate the NeRF to get a collection of 3D quantities such as the surface normal (and others to be defined shortly). We then alpha composite these quantities to get the expected surface termination point $\avgx$, and the ray's associated surface normal $\avgnormal$,
by applying the volume rendering typically used for color $\{\mathbf{c}^{(i)}\}_{i=1}^N$ in Equation~\ref{eq:nerfcolor} to the sample points  $\{\vx^{(i)}\}_{i=1}^N$ and normals $\{\normal^{(i)}\}_{i=1}^N$:
\begin{equation} \label{eq:volrenderxn}
    \avgx = \sum_{i=1}^N w^{(i)} \vx^{(i)}\,,\quad\quad
    \avgnormal = \sum_{i=1}^N w^{(i)} \normal^{(i)}\,.
\end{equation}

We then construct a new reflected ray with
direction $\refdir$ by reflecting the initial ray about the surface normal:
\begin{equation} \label{eq:meanrefray}
    \refdir = \viewdir - 2(\bar{\normal}\cdot\viewdir)\bar{\normal}\,.
\end{equation}

Tracing a single reflection ray cannot be used to render materials that are not perfect mirrors, and it ignores the fact that Zip-NeRF traces cones instead of rays for each pixel. Accordingly, we model a cone-like distribution of reflected rays. The shape of this distribution is affected by two factors: the radius of the original pixel cone cast from the camera, and the roughness of the surface at the expected termination point of the camera ray, as visualized in Figure~\ref{fig:refrays}. We start by modeling both the cone-like distribution of reflected rays and the distribution of reflection directions due to surface roughness as von Mises-Fisher (vMF, \ie a normalized spherical Gaussian) distributions centered at the mirror reflection direction $\refdir$. The vMF width (reciprocal of its concentration parameter $\kappa$) for the pixel cone distribution is the pixel radius $\dot{r}$, and the vMF width for the surface roughness is a quantity $\roughness$ that we parameterize at any point in space. The reflected ray cone at the camera ray's expected termination point is modeled as a vMF distribution $\operatorname{vMF}(\refdir, \avgkappa)$, whose width $\bar{\kappa}^{-1}$ is the sum of the pixel ray cone's radius $\dot{r}$ and the composited roughness $\bar{\roughness}$ along the primary ray:
\begin{equation}
    \bar{\kappa}^{-1} = \dot{r} + \bar{\roughness}\,.
\end{equation}

In order to ensure that the cross-section of the reflected ray cone at the surface matches that of the incident ray, we apply a roughness-dependent shift to the reflected ray cone's origin so that it intersects $\avgx$ while having the incident ray's radius $\dot{r}\|\vo-\avgx\|$ (see Figure~\ref{fig:refrays}(c) and Appendix~\ref{app:reflected_ray} for more details):
\begin{equation} \label{eq:shiftedorigin}
    \vo' = \avgx - \lft\|\vo-\avgx\rgt\|\frac{\dot{r}}{\dot{r}+\bar{\roughness}} \refdir\,.
\end{equation}
Note that when the accumulated roughness $\bar{\roughness}$ is zero, $\vo'$ is not shifted towards the surface, and is instead perfectly mirrored to the other side of the surface.

\subsection{Conical Reflected Features}
\label{sec:aa}\label{sec:generalenc}

Now that we have defined a vMF distribution over reflected rays, our goal is to estimate the expected volume-rendered feature over the vMF distribution, which we can then decode to a reflected color. This expected feature can be written as:
\begin{equation} \label{eq:vmfconcentration}
    \avgf^\star = \mathbb{E}_{\vomega\sim\operatorname{vMF}(\refdir,\avgkappa)}\lft[\avgf(\vomega)\rgt]=\int_{\mathbb{S}^2} \avgf(\vomega) \operatorname{vMF}\lft(\vomega ;\refdir, \bar{\kappa}\rgt)\, d\vomega\,,
\end{equation}
where $\avgf(\vomega)$ is the feature vector accumulated in the direction $\vomega$. 

Estimating this integral using Monte Carlo over randomly-sampled rays is prohibitively expensive since each sample requires volumetric rendering along the ray. Inspired by Zip-NeRF, we approximate this integral using a small set of representative samples combined with feature downweighting. However, unlike Zip-NeRF, we perform both of these operations in the 2D directional domain rather than in 3D Euclidean space.

\paragraph{Directional Sampling}

Instead of sampling a large number of random reflection rays, we select a representative set of rays using unscented directional sampling~\cite{kurz2016unscented}. We evaluate $K$ rays $\{(\vo'_j, \refdir_j)\}_{j=1}^{K}$ (in our experiments we set $K=5$) selected to maintain the mean $\refdir$ and concentration $\bar{\kappa}$, by setting $\refdir_1 = \refdir$ and $\{\refdir_j\}_{j=2}^{K}$ placed on a circle around $\refdir$ whose radius depends on $\avgkappa$. During optimization we randomly rotate the samples rigidly on the circle and during evaluation we fix them. A more detailed description of this procedure is provided in Appendix~\ref{app:directional_sampling}.

For each of the $K$ sampled rays $(\vo'_j, \refdir_j)$, we evaluate the volume density and appearance features at $N'$ sample points along the ray, and volume render them into the feature $\avgf(\refdir_j)$. In order to generate these $N'$ samples we use the same procedure as sampling the $N$ rays on the camera rays (see supplement for more details). The $K$ features are then averaged to yield a single feature vector for the reflected cone:
\begin{equation} \label{eq:reffeatures}
    \avgf(\refdir_j) = \sum_{i=1}^{N'} w_j^{(i)} \vf\lft(\vx_j^{(i)}\rgt)\,, \quad\quad \avgf = \frac{1}{K}\sum_{j=1}^{K} \avgf\lft(\refdir_j\rgt)\,,
\end{equation}
where $w_j^{(i)}$ and $\vx_j^{(i)}$ are the $j$th reflected ray's $i$th sample weight and 3D location, respectively, and $\vf(\vx)$ is the feature at $\vx$.

\paragraph{Reflection Feature Downweighting}

The directional sampling described above helps select a small  representative set of rays to average. However, for surfaces with high roughness, the sampled rays can be distant from one another relative to the underlying 3D grid cells. This means that the feature from Equation~\ref{eq:reffeatures} may be aliased, and small variations in the reflected ray directions can cause large variations in appearance. 

In order to prevent this, we adapt the ``feature downweighting'' technique from Zip-NeRF to our directional setting. We do this by multiplying features corresponding to voxels that are small relative to the vMF cone by a small multiplier, reducing their effect on the rendering color.
Following Zip-NeRF, we define the downweighted feature at a point $\vx$ as:
\begin{equation}
    \vf_{\text{aa}}(\vx) = \operatorname{erf}\lft(\lft(\sqrt{8}\boldsymbol{\nu} \sigma(\vx) \rgt)^{-1}\rgt)\odot\vf(\vx)\,,
\end{equation}
where $\sigma(\vx)$ is the (scalar) scale of the cone at the point $\vx$ (defined below), $\boldsymbol{\nu}$ is a vector of the same dimension as $\vf$ containing the scale of NGP grid resolution corresponding to every entry of $\vf$, $\odot$ denotes elementwise multiplication, and the reciprocal and $\operatorname{erf}(\cdot)$ are taken elementwise. The downweighted feature $\vf_{\text{aa}}(\vx)$ is used in Equation~\ref{eq:reffeatures} in place of the original feature $\vf(\vx)$.

Since we focus on large scenes, which require using a contraction function $\mathcal{C}$ (Equation~\ref{eq:contraction}), it is particularly important to consider the behavior of $\sigma(\vx)$ in the nonlinear region of the contraction. Zip-NeRF scales $\sigma(\vx)$ by the geometric mean of the contraction function's 3D Jacobian's eigenvalues (Equation~\ref{eq:zipnerfscale}). This has the unfortunate effect of causing $\sigma(\vx)$ to approach 0 (\ie, applying no downweighting) for points $\vx$ far from the origin, because $\mathcal{C}$ strongly contracts distant points towards the origin. As a result, using Zip-NeRF's downweighting function would result in significant aliasing in reflections of distant content. We instead use the Jacobian determinant restricted to the (2D) directional domain, and define:
\begin{equation} \label{eq:scale}
    \sigma(\vx) = \scalemult\cdot(\dot{r}+\bar{\roughness})\lft\|\vx - \vo'\rgt\| \sqrt{\det\jacobian_\mathcal{C}^{\operatorname{dir}}(\vx)}\,,
\end{equation}
where $\scalemult$ is a fixed scale multiplier which we set to $16$ in all experiments, and:
\begin{align}
    \det\jacobian_\mathcal{C}^{\operatorname{dir}}(\vx) &= \det\jacobian_\mathcal{C}(\vx)\cdot\lft(\frac{\partial}{\partial \|\vx\|}\lft(\mathcal{C}(\vx)\cdot\frac{\vx}{\|\vx\|}\rgt)\rgt)^{-1} \\
    &= \lft(\frac{2\max(1, \|\vx\|) - 1}{\phantom{2}\max(1, \|\vx\|)^2 \phantom{-1}}\rgt)^2\,.
\end{align}
Note that when the point $\vx$ is far from the origin, the scale in Equation~\ref{eq:scale} approaches a constant, $\sigma \xrightarrow[]{\|\vx\|\rightarrow\infty} 2\scalemult(\dot{r}+\bar{\roughness})$, in contrast to Zip-NeRF's scale, which approaches zero and therefore does not downweight distant content. While this may be less noticeable in Zip-NeRF, properly downweighting distant content is crucial for us to prevent aliasing in reflections of distant illumination.

\subsection{Color Decoder} \label{sec:appearance}

The role of the color decoder is to assign a color to every sample point along a ray. Inspired by UniSDF~\cite{wang2023unisdf}, our color decoder uses a convex combination of two color components:
\begin{equation}
    \rgb\lft(\vx^{(i)}, \viewdir\rgt) = \beta^{(i)}\rgb_v\lft(\vx^{(i)}, \viewdir\rgt) + \lft(1-\beta^{(i)}\rgt)\rgb_r\lft(\vx^{(i)}, \refdir\rgt)\,,
\end{equation}
where $\beta^{(i)} \in [0, 1]$ is a weighting coefficient output by the geometry encoder followed by a sigmoid nonlinearity. The first color component $\rgb_v$ is similar to the typical NeRF view-dependent appearance model:
\begin{equation} \label{eq:rgbv}
    \rgb_v\lft(\vx^{(i)}, \viewdir\rgt) = g\lft(\vx^{(i)}, \vb^{(i)}, \normal^{(i)}, \viewdir \rgt)\,,
\end{equation}
and the second component $\rgb_r$, designed to model glossy appearance, is computed as:
\begin{equation} \label{eq:rgbr}
    \rgb_r\lft(\vx^{(i)}, \refdir\rgt) = h\lft(\vx^{(i)}, \vb^{(i)}, \normal^{(i)},  \viewdir\cdot\normal^{(i)}, \refdir, \avgf\rgt)\,,
\end{equation}
where $g$ and $h$ are small MLPs (see supplement for more details), and $\avgf$ is the reflection feature introduced in Section~\ref{sec:generalenc}. The bottleneck vector $\vb$, similar to $\beta$, is also output by a small MLP applied to geometry features. Note that unlike UniSDF, we apply $\beta$ in 3D space rather than in image space, and therefore the reflected features $\avgf$ are fed into the color decoder at every sample along the camera ray. We find that this is useful for convergence since early in optimization, when geometry is still fuzzy, every point along the ray can make use of the reflected feature. See full description of the color decoder in the supplement.

\subsection{Geometry Representation and Regularization} \label{sec:geometry}

Our geometry representation is based on that of Zip-NeRF~\cite{barron2023zipnerf}. See its full description in Appendix~\ref{sec:geometrysupp}. Like Ref-NeRF~\cite{verbin2022refnerf}, we apply orientation loss to the normals, and we make use of predicted normals $\tilde{\normal}$ output by a small MLP applied to the NGP features. Unlike Ref-NeRF, we use an \emph{asymmetric} predicted normal loss which allows separate multipliers for the gradients flowing from the geometry normals $\normal$ (\ie the normalized negative gradient of density) and rendering weights $w$ along the ray to the predicted normals $\tilde{\normal}$, and vice versa:
\begin{align} \label{eq:prednormalloss}
    &\mathcal{L}_{\operatorname{pred}} = \lambda_n \mathcal{L}_p\lft(w, \normal, \stopgrad\lft(\tilde{\normal}\rgt)\rgt) + \lambda_{\tilde{n}} \mathcal{L}_p\lft(\stopgrad\lft(w\rgt), \stopgrad\lft(\normal\rgt), \tilde{\normal}\rgt)\,,\\
    &\text{where }\; \mathcal{L}_p(w, \normal, \tilde{\normal}) =\sum_{i=1}^N w^{(i)} \lft\|\normal^{(i)} - \tilde{\normal}^{(i)}\rgt\|^2.
\end{align}
Where $\stopgrad(\cdot)$ is a stop-gradient operator, which prevents gradients from flowing to its input.

In all of our experiments we use the asymmetric predicted normal loss with $\lambda_n = 10^{-3}$ and $\lambda_{\tilde{n}} = 0.3$. This strongly ties the predicted normals to the ones corresponding to the NeRF's density, while not oversmoothing geometry. This allows the predicted normals to resemble the geometry normals while being significantly smoother, which makes them useful for computing the reflection directions and creating accurate specular highlights.

\newcommand{\ablationwidth}{0.49\linewidth}

\begin{figure}
    \centering
    \begin{tabular}{@{}c@{\,\,}c@{}}
        \includegraphics[width=\ablationwidth]{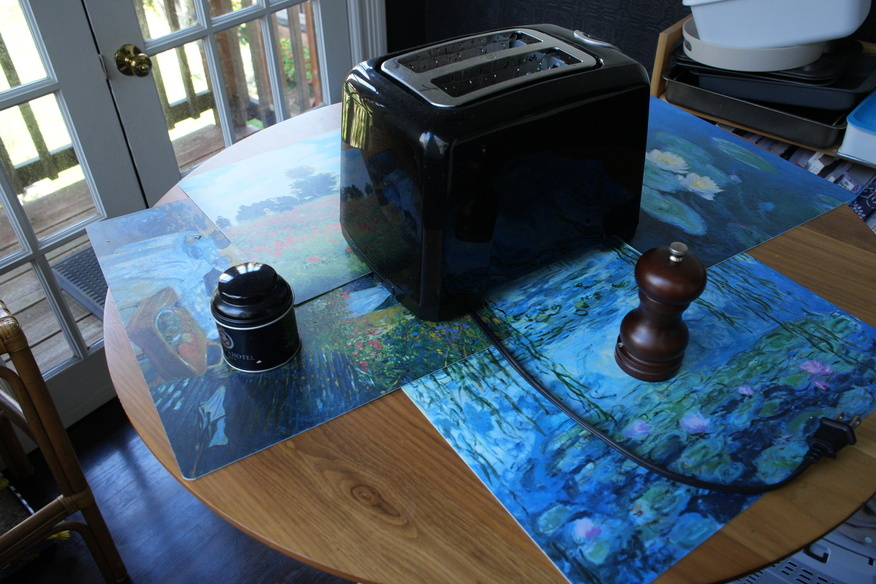} &
        \includegraphics[width=\ablationwidth]{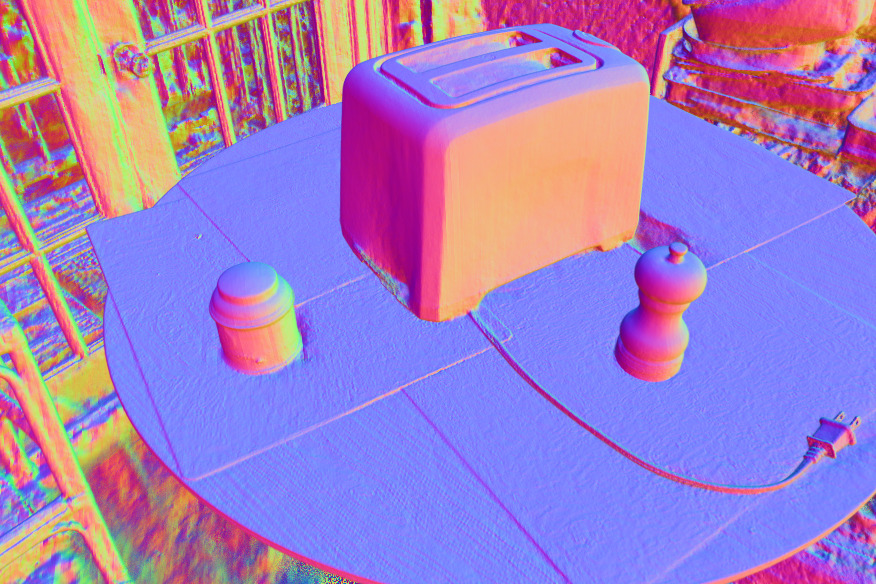}\vspace{-0.1cm} \\
        \footnotesize Input image & \footnotesize (a) Our recovered surface normals \\[3pt]
        \includegraphics[width=\ablationwidth]{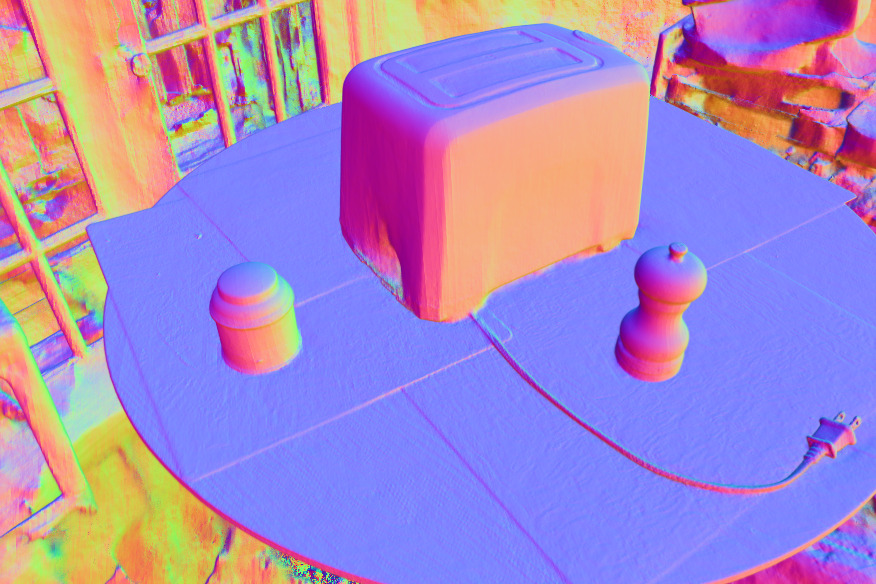} &
        \includegraphics[width=\ablationwidth]{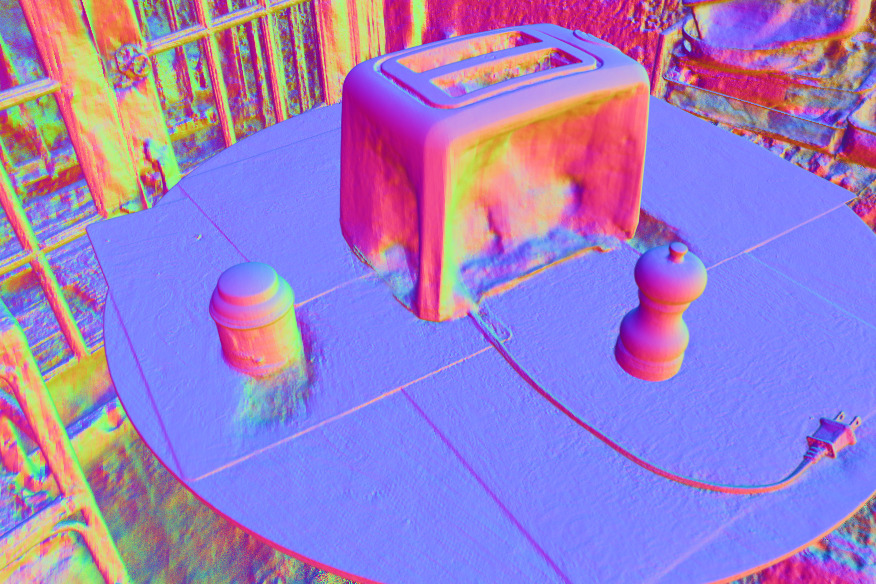}\vspace{-0.1cm} \\
        \footnotesize (b) Symmetric loss $\lambda_n=\lambda_{\tilde{n}}=10^{-2}$ & \footnotesize (c) Symmetric loss $\lambda_n=\lambda_{\tilde{n}}=10^{-3}$ \\[3pt]
        \includegraphics[width=\ablationwidth]{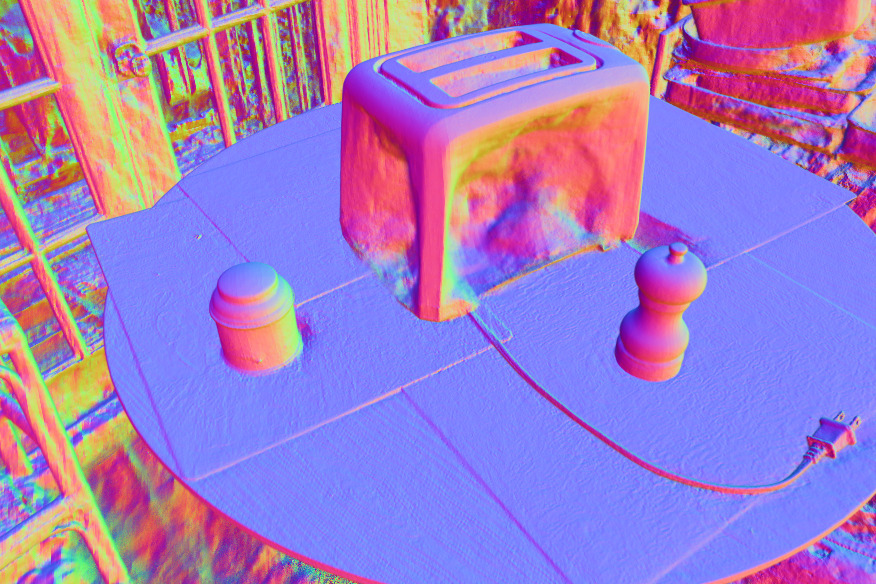} &
        \includegraphics[width=\ablationwidth]{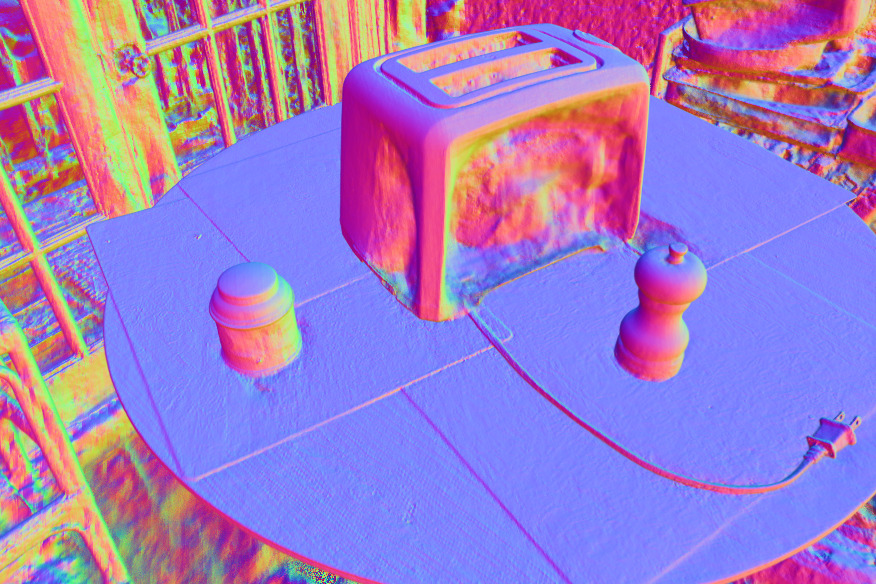}\vspace{-0.1cm} \\
        \footnotesize (d) Shared reflection features & \footnotesize (e) Single dilated cone
    \end{tabular}
    \vspace{-0.15in}
    \caption{An ablation study showing the effect of different design choices on our recovered normals. Replacing (a) our model's asymmetric predicted normal loss (Equation~\ref{eq:prednormalloss}) with the standard symmetric one tends to (b) oversmooth or (c) underconstrain geometry, preventing the normal vectors from converging to the correct solution. (d) Using reflected features $\avgf$ which are also used for other appearance components such as the bottleneck vector $\vb$, or (e) using a single cone instead of $K$ directional sampling cones also often result in poor recovery of surface normals.
    \label{fig:normals}
    }
\end{figure}

\section{Experiments and Results}

We evaluate our method on four datasets. First, we use the three real scenes from Ref-NeRF~\cite{verbin2022refnerf} to demonstrate that our method decisively outperforms baselines for reconstructing and rendering highly reflective objects. Second, we use the nine scenes from mip-NeRF 360~\cite{barron2022mipnerf360} to show that our method's performance is not limited to shiny scenes and that we are on par with state-of-the-art techniques for rendering novel views of large-scale mostly-diffuse scenes. Third, we use four newly-captured scenes to demonstrate specific effects and capabilities of our method, such as its ability to accurately capture near-field reflections. Finally, we show results on the synthetic \emph{Shiny Blender} dataset from Ref-NeRF~\cite{verbin2022refnerf}.

Throughout this section we provide quantitative and qualitative results showing the benefit of our methods and the effects of its components. However, image error metrics are frequently dominated by effects prevalent in real data, such as camera miscalibration, and illumination and appearance changes, and because the smooth motion of the reflections is a key benefit of our method, we supplement our paper with an interactive project page containing many video results demonstrating these effects, and encourage the reader to view it.

We implement our method in JAX~\cite{jax2018github}. Optimization is done using $8$ V100 GPUs, which takes our method approximately $100$ minutes to optimize for a single scene, compared with $50$ minutes for Zip-NeRF and $200$ minutes for UniSDF. All of our experiments are performed using the same set of hyperparameters (see supplement).

\begin{table}[!]
    \centering
\resizebox{1\linewidth}{!}{
    \begin{tabular}{l@{\,\,}|ccc}
    & \!PSNR$\uparrow$\! & \!SSIM$\uparrow$\! & \!LPIPS$\downarrow$\ \\ \hline
UniSDF & 33.38 & 0.961 & 0.042 \\
Zip-NeRF & 33.09 & 0.960 & 0.041 \\
Ref-NeRF* & 32.59 & 0.957 & 0.042 \\\hline
 Ours with single downweighted cone & 33.63 & \cellcolor{orange} 0.964 & \cellcolor{orange} 0.038 \\
Ours with single dilated cone & \cellcolor{orange} 33.72 & \cellcolor{orange} 0.964 & \cellcolor{orange} 0.038 \\
Ours without downweighting & 33.63 & \cellcolor{orange} 0.964 & \cellcolor{orange} 0.038 \\
Ours with 3D Jacobian & \cellcolor{yellow} 33.65 & \cellcolor{yellow} 0.963 & \cellcolor{yellow} 0.039 \\
Ours without near-field reflections & 33.30 & 0.962 & 0.040 \\
Ours & \cellcolor{red} 33.91 & \cellcolor{red} 0.965 & \cellcolor{red} 0.036
    \end{tabular}
    }
    \caption{Average results across the three real scenes from Ref-NeRF~\cite{verbin2022refnerf} and four new real scenes we captured ourselves. Metrics only report performance on shiny image regions. The ``Ref-NeRF*'' baseline is an improved version of Ref-NeRF that uses Zip-NeRF's geometry model, sampling and optimization procedure, but with Ref-NeRF's appearance model. See the supplement for many more quantitative results.
    }
    \label{tab:mainresults}
\end{table}

\subsection{Evaluation Details}

We evaluate all results using the three metrics commonly used for view synthesis: PSNR, SSIM~\cite{wang2004image}, and LPIPS~\cite{zhang2018lpips}. In addition to these metrics, which are computed over entire images, we compute ``masked'' PSNR, SSIM, and LPIPS. These masked metrics are computed by compositing the shiny regions (in both the rendered and ground truth images) onto a white background, and then computing the standard metrics. We compute the masked metrics for all of our newly-captured scenes as well as the three scenes from the Ref-NeRF real dataset. The masked metrics are used to highlight the benefit of our method by focusing on shiny regions of the captured scenes. See supplement for examples of these masks.

\subsection{Baseline Comparisons}

We compare our method with prior work designed for photorealistic view synthesis for general scenes without significant reflective surfaces~\cite{barron2023zipnerf,kerbl3Dgaussians}, as well as methods specially-designed for scenes with shiny objects~\cite{verbin2022refnerf,liang2023envidr,wang2023unisdf}, as well as concurrent work such as NDE~\cite{liwen} and SpecNeRF~\cite{specnerf}. All baseline experiments were done with the gracious help of the original authors, or otherwise using the official codebases. 

Table~\ref{tab:mainresults} shows a comparison of our results with prior work, using the masked metrics, showing that our method is indeed more accurate than prior work in regions featuring shiny surfaces. This is further exemplified qualitatively by our results in Figures~\ref{fig:teaser},~\ref{fig:normals},~\ref{fig:sphere_ablation},~\ref{fig:nearfield},~\ref{fig:spheres_comp}, and~\ref{fig:toaster_comp}. Our supplemental webpage also shows an interactive comparison of our results with those obtained by prior work.

Our supplemental tables also show per-scene breakdowns of the metrics in Table~\ref{tab:mainresults}, as well as their unmasked counterparts, metrics on the standard mip-NeRF 360 dataset~\cite{barron2022mipnerf360} featuring mostly-diffuse scenes, and the Shiny Blender dataset from Ref-NeRF~\cite{verbin2022refnerf}, on top of two additional shiny scenes from the standard Blender dataset~\cite{mildenhall2020nerf}).

\newcommand{\ablatewidth}{0.195\linewidth}
\begin{figure*}
    \begin{tabular}{@{}c@{\,\,}c@{\,\,}c@{\,\,}c@{\,\,}c@{}}
        \includegraphics[width=\ablatewidth, trim={280px 50px 300px 240px}, clip]{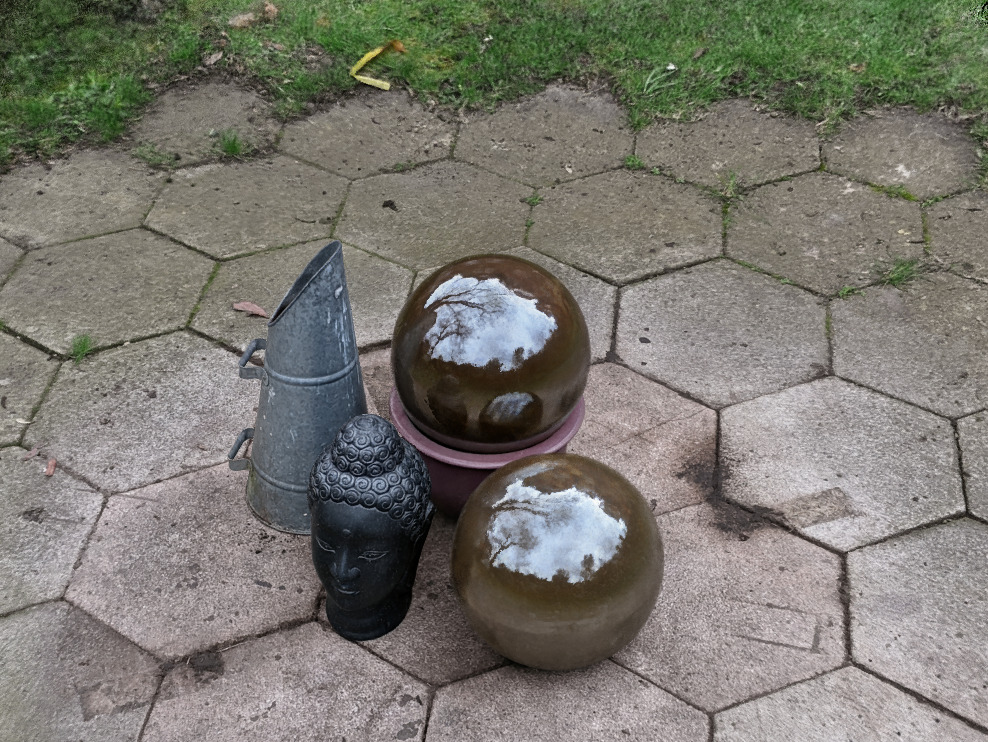} &
        \includegraphics[width=\ablatewidth, trim={280px 50px 300px 240px}, clip]{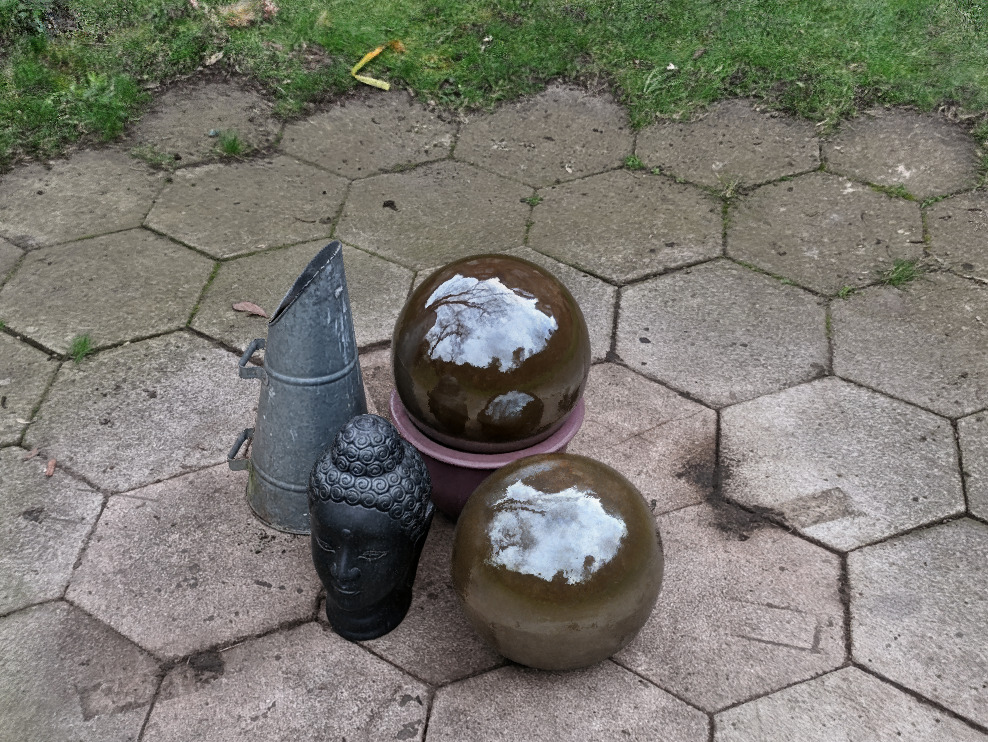} &
        \includegraphics[width=\ablatewidth, trim={280px 50px 300px 240px}, clip]{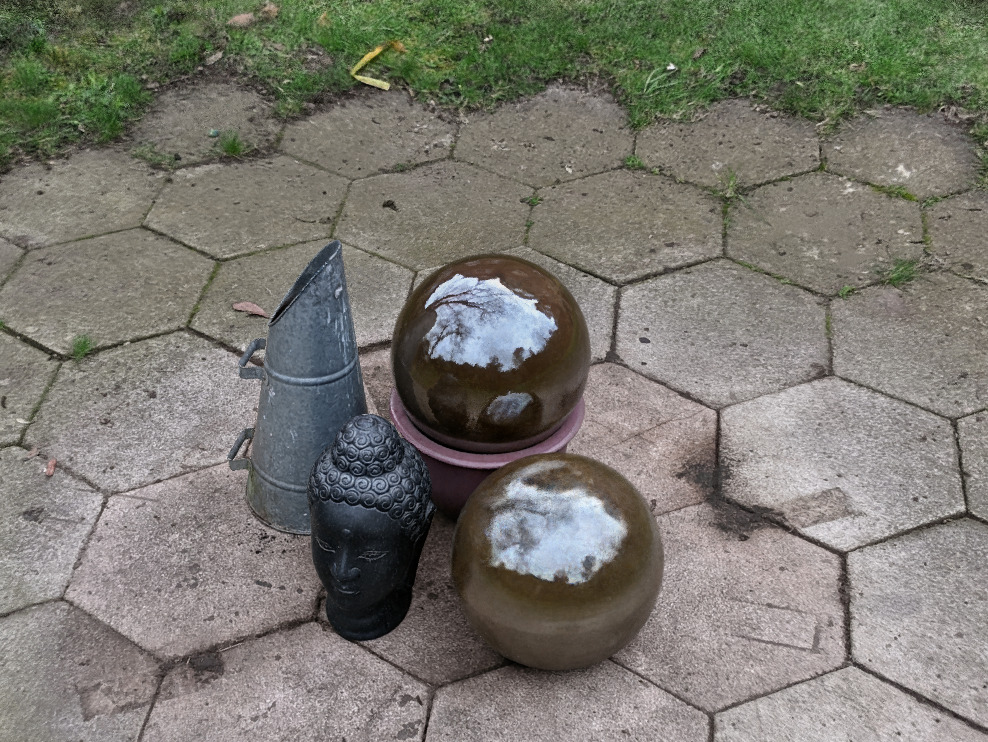} &
        \includegraphics[width=\ablatewidth, trim={280px 50px 300px 240px}, clip]{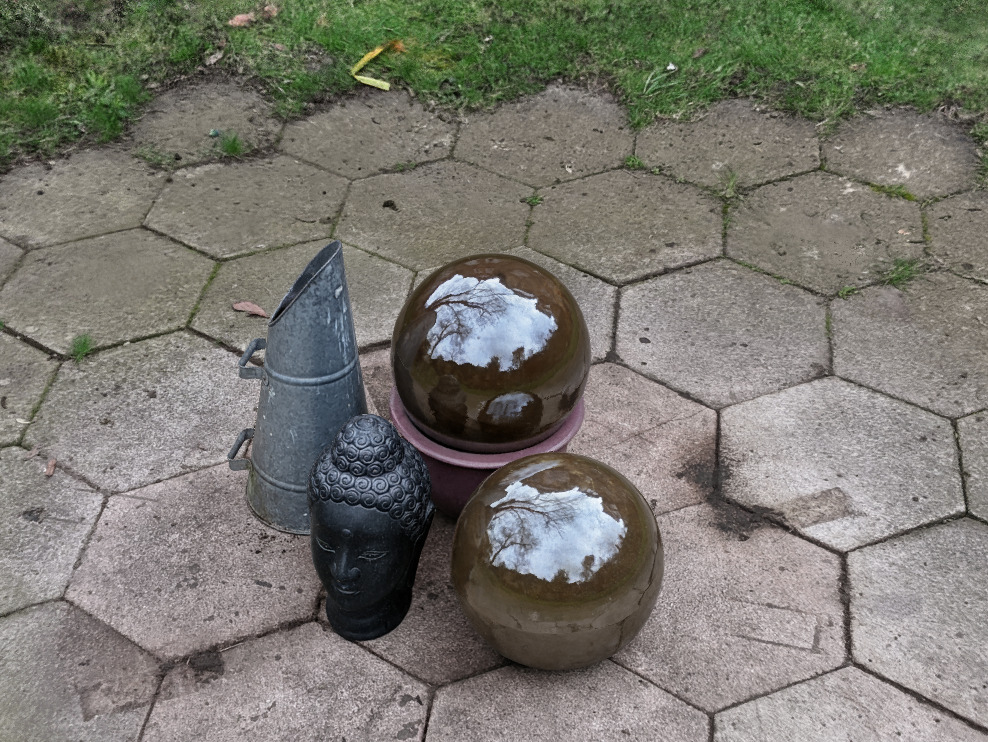} &
        \includegraphics[width=\ablatewidth, trim={280px 50px 300px 240px}, clip]{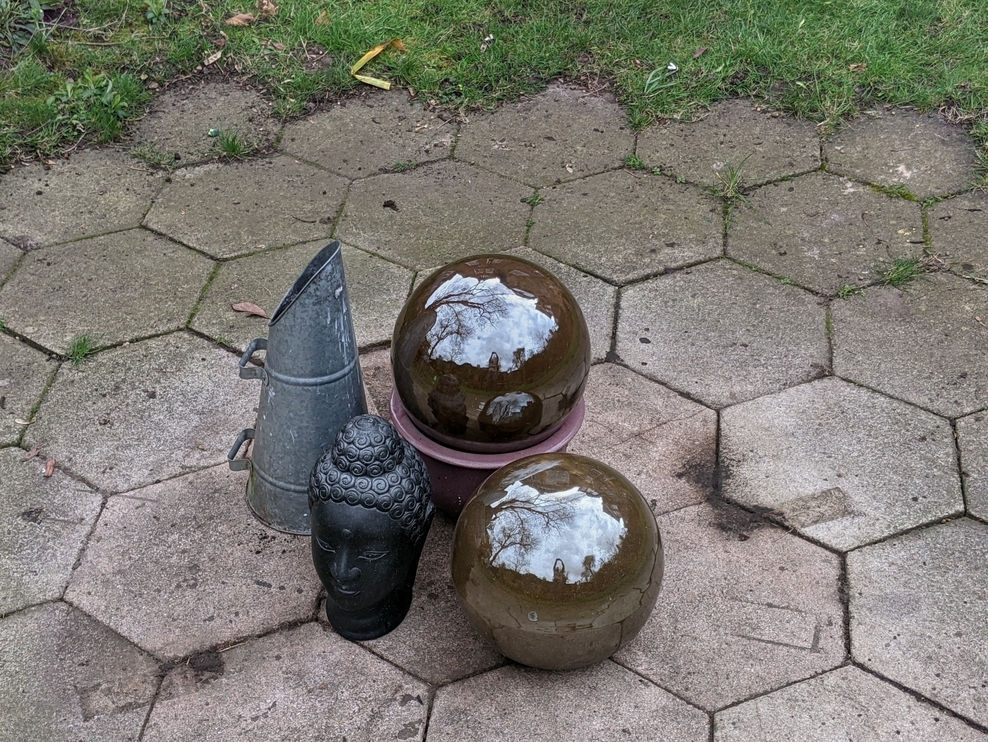} 
        \\
        \small (a) Single reflection ray & \small (b) No downweighting & \small (c) 3D Jacobian & \small (d) Ours & \small Ground truth
    \end{tabular}
    \vspace{-0.13in}
    \caption{Ablation of our reflection anti-aliasing components. Using (a) a single reflection ray instead of five, (b) not downweighting reflection features, or (c) using Zip-NeRF's 3D Jacobian instead of restricting it to the 2D directional Jacobian, all result in inaccurate reflections. Even for low-roughness objects such as the shiny spheres shown here, aliasing in the reflections during optimization prevents the model from accurately reconstructing both the reflective surface geometry as well as the reflected content.}
    \label{fig:sphere_ablation}
\end{figure*}

\subsection{Ablation Studies}

In this section we carefully ablate the key components of our method to justify our design decisions. Table~\ref{tab:mainresults}, and Tables~\ref{tab:ablationmasked} and~\ref{tab:ablation} in the supplement contain quantitative results for these ablation studies, and the supplemental webpage contains video comparisons for them.

\paragraph{Tracing cones through the recovered scene synthesizes more accurate reflections.}

Figure~\ref{fig:nearfield} shows that replacing the cone tracing operation with simply gathering reflection features at infinity (which depends only on the reflection direction) still recovers far-field reflections, but cannot accurately reproduce near-field reflections.

\paragraph{Asymmetric predicted normal loss, separate reflection features, and multiple cones are all crucial for recovering accurate geometry.}

Figure~\ref{fig:normals}(b,c) demonstrates the importance of our asymmetric normal loss (Equation~\ref{eq:prednormalloss}). Using Ref-NeRF's normal loss can oversmooth geometry when its weight is too high. It can also result in incorrect geometry when turned too low, since the predicted normals can greatly deviate from the ones corresponding to geometry, which lets them overfit to each image's specular highlights without properly generalizing to unseen views.

Figure~\ref{fig:normals}(d) shows that using the same NGP components for both the reflected features $\avgf$ and the appearance bottleneck vector $\vb$ may also result in poorly-recovered geometry, due to the strong dependence of the reflection features on the surface normals. Additionally, in Figure~\ref{fig:normals}(e) we show that using a single Zip-NeRF cone with radius $\dot{r}' = \dot{r}+\bar{\roughness}$ instead of our directional sampling can lead to inaccurate geometry. This is because the model is only supervised at cone radii corresponding to the pixel footprints $\dot{r}$, and therefore evaluating it with dilated radii $\dot{r}'$ may not correspond to the true underlying geometry, adversely affecting reconstruction.

\paragraph{Anti-aliasing methods improve recovered geometry and reflections.}

Figure~\ref{fig:sphere_ablation} visualizes the impact of our reflection anti-aliasing strategies on rendered specularities. Using only a single reflection ray, omitting the reflection feature downweighting, or using Zip-NeRF's 3D Jacobian instead of our 2D directional Jacobian for downweighting, all decrease the accuracy of rendered reflections. Intuitively, if reflections are aliased during optimization, the model is unable to recover accurate surface geometry or accurate geometry and features for the reflected content. 

\subsection{Discussion}

Our method quantitatively outperforms existing view synthesis techniques, particularly for shiny surfaces that display detailed specular reflections. However, we believe that the qualitative visual improvement substantially outweighs the quantitative improvement in image metrics and we strongly urge readers to view our supplementary videos. Most notably, the smooth and consistent motion of reflections synthesized by our method is strikingly more realistic than the view-dependent appearance rendered by baseline methods. This suggests that standard image error metrics (PSNR, SSIM, etc.) are not adequate for evaluating the quality of view-dependent appearance.

\subsection{Limitations}

While our method is successful in its goal of accurately reconstructing and rendering reflective surfaces, it is not able to fully model certain effects. Because we only reflect from the expected termination point of each camera ray, our method struggles to render semi-transparent surfaces. Additionally, the camera operator is frequently visible in reflective surfaces and our model does not consider this potential source of error.

\section{Conclusion}

We have presented a method for rendering scenes containing highly specular objects, by using a neural radiance field. Our method relies on reflecting cones off of surfaces in the scene and tracing them through the NeRF, and on a novel set of techniques designed to anti-alias these reflections, and is able to synthesize accurate detailed reflections of both distant and near-field content that consistently and smoothly moves across surfaces. We demonstrate that our approach quantitatively outperforms prior view synthesis models, and that it provides great qualitative improvement over prior work's realism in rendering novel views.

\begin{acks}
We wish to thank Georgios Kopanas, Christian Richardt, Li Ma, and Liwen Wu for helping with comparison results and figures. We would also like to thank Cassidy Curtis, Janne Kontkanen, Aleksander Hołyński, Alex Trevithick, and Georgios Kopanas for ideas and comments on our work.
\end{acks}

\bibliographystyle{ACM-Reference-Format}
\bibliography{bibliography}

\cleardoublepage

\newcommand{\nearfieldwidth}{0.245\linewidth}
\begin{figure*}
    \centering
    \begin{tabular}{@{}c@{\,\,}c@{\,\,}c@{\,\,}c@{}}
        \includegraphics[width=\nearfieldwidth, trim={120mm 90mm 30mm 5mm}, clip]{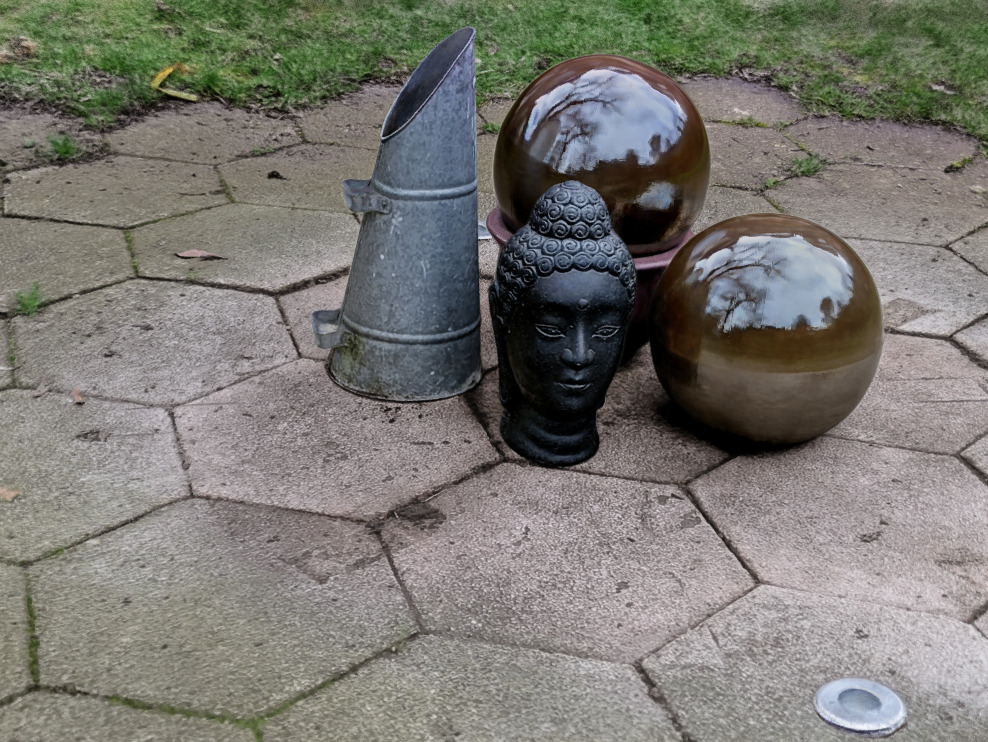} &
        \includegraphics[width=\nearfieldwidth, trim={120mm 90mm 30mm 5mm}, clip]{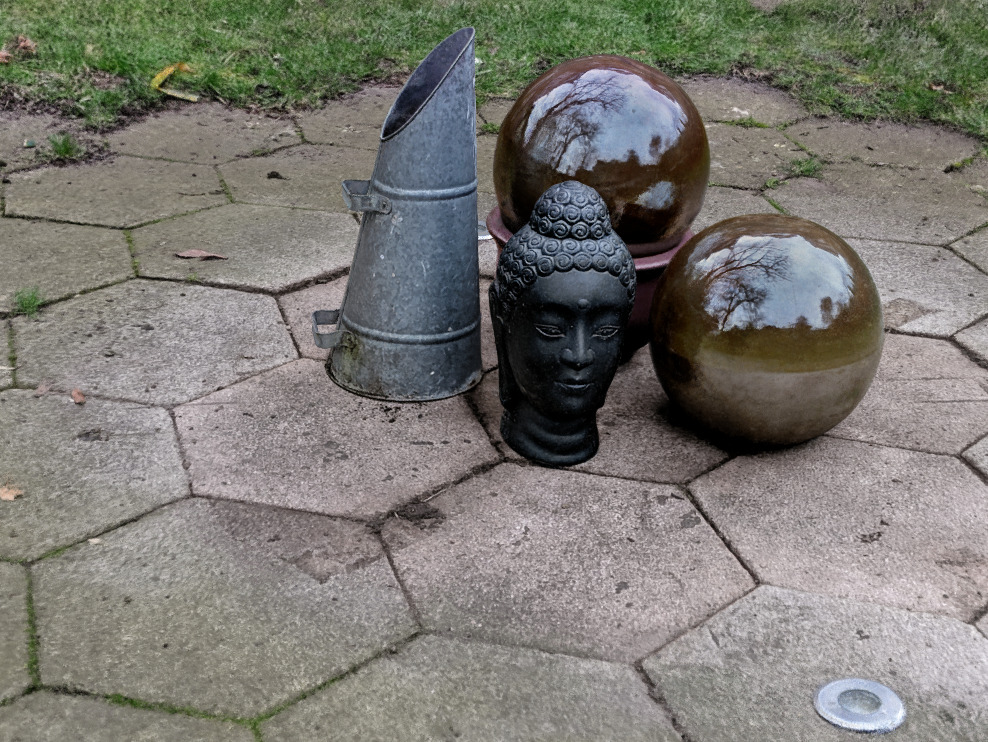} &
        \includegraphics[width=\nearfieldwidth, trim={120mm 90mm 30mm 5mm}, clip]{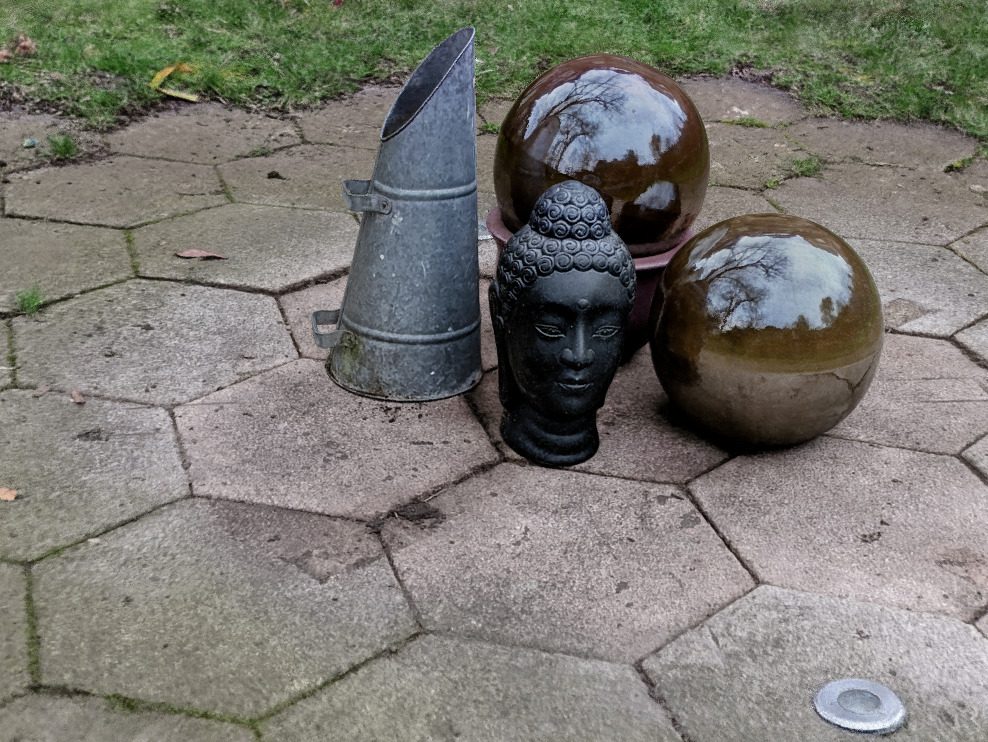} & \includegraphics[width=\nearfieldwidth, trim={120mm 90mm 30mm 5mm}, clip]{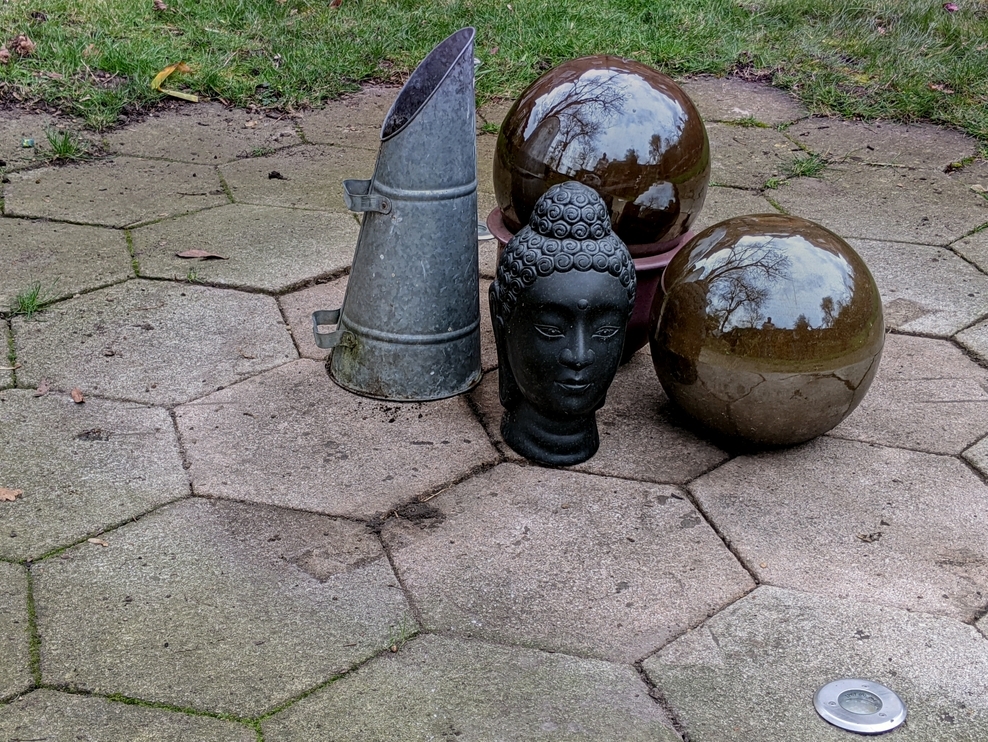}
        \\ 
        \includegraphics[width=\nearfieldwidth, trim={290px 193px 193px 97px}, clip]{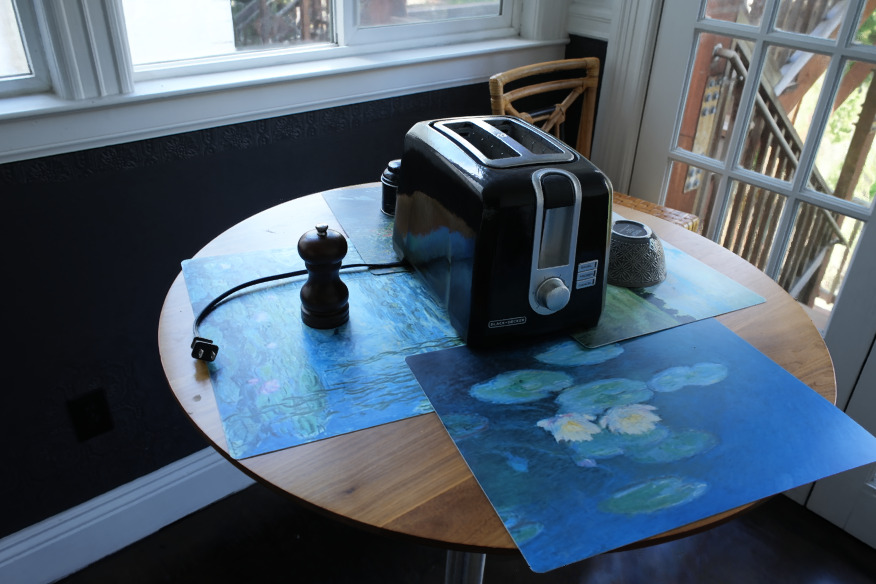} &
        \includegraphics[width=\nearfieldwidth, trim={290px 193px 193px 97px}, clip]{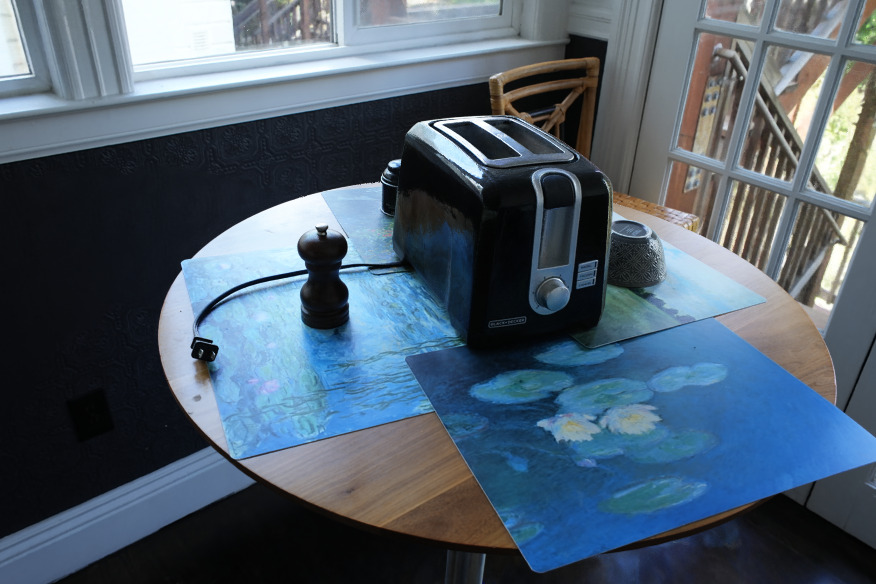} &
        \includegraphics[width=\nearfieldwidth, trim={290px 193px 193px 97px}, clip]{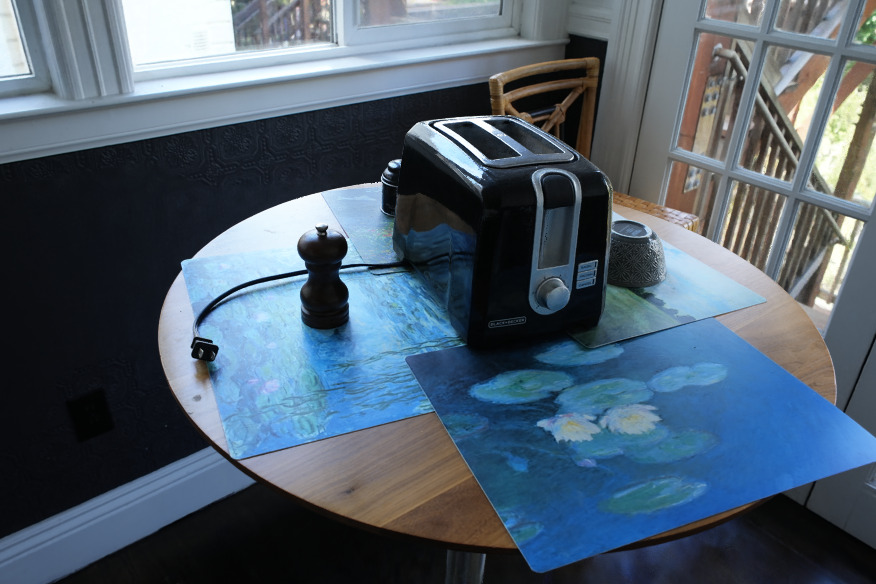} &         \includegraphics[width=\nearfieldwidth, trim={60px 40px 40px 20px}, clip]{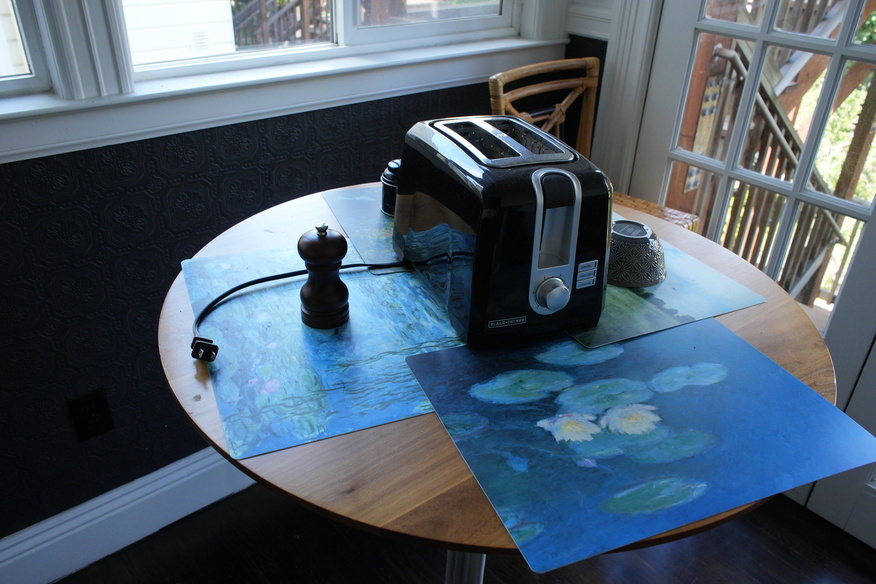} \\
        \small (a) UniSDF & \small (b) Ours without traced cones & \small (c) Ours & \small Ground truth
    \end{tabular}
    \vspace{-0.15in}
    \caption{Two examples showing that (a) tracing cones and intersecting them with the geometry of the NeRF allows our model to recover and render near-field reflections such as the statue head and cone reflected in the balls, and the cord and artwork reflected in the toaster. This is in contrast to (b) our model with the feature grid only queried infinitely-far away, or (c) UniSDF, both of which only use the reflection direction. Note that both our model in (a) and our ablated model in (b) are capable of rendering accurate far-field reflections, while UniSDF renders blurry reflections even for far-field content.
    \label{fig:nearfield}
    }
\end{figure*}

\newcommand{\spherecompwidth}{0.16\linewidth}
\begin{figure*}
    \centering
    \begin{tabular}{@{}c@{\,\,}c@{\,\,}c@{\,\,}c@{\,\,}c@{\,\,}c@{}}
        \includegraphics[width=\spherecompwidth, trim={125mm 120mm 75mm 30mm}, clip]{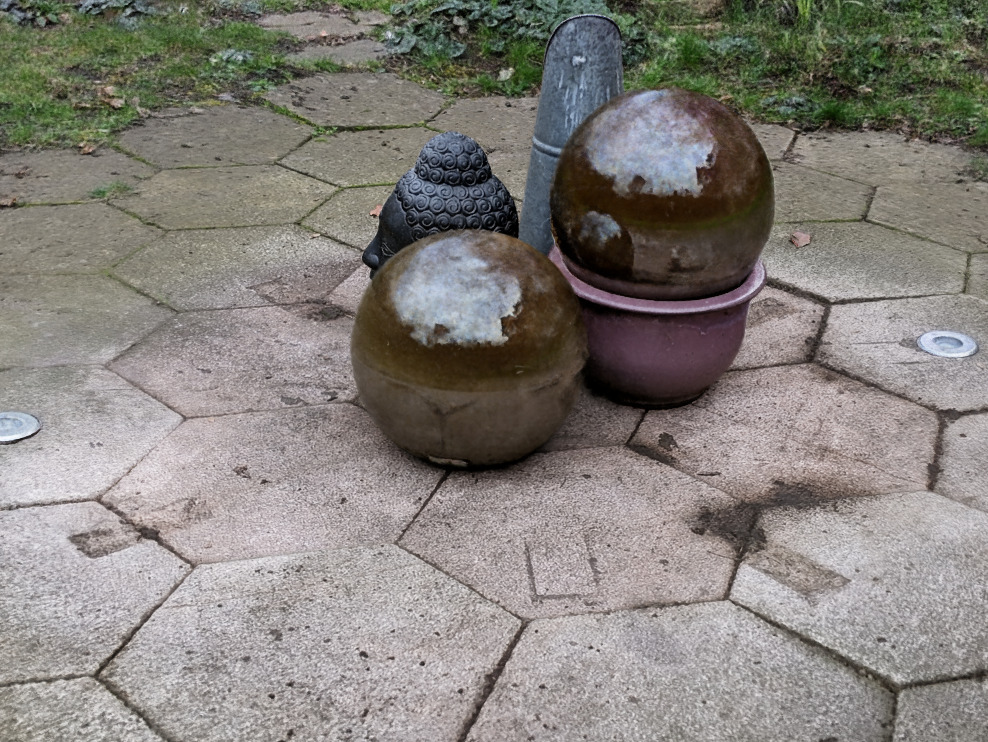} &
        \includegraphics[width=\spherecompwidth, trim={125mm 120mm 75mm 30mm}, clip]{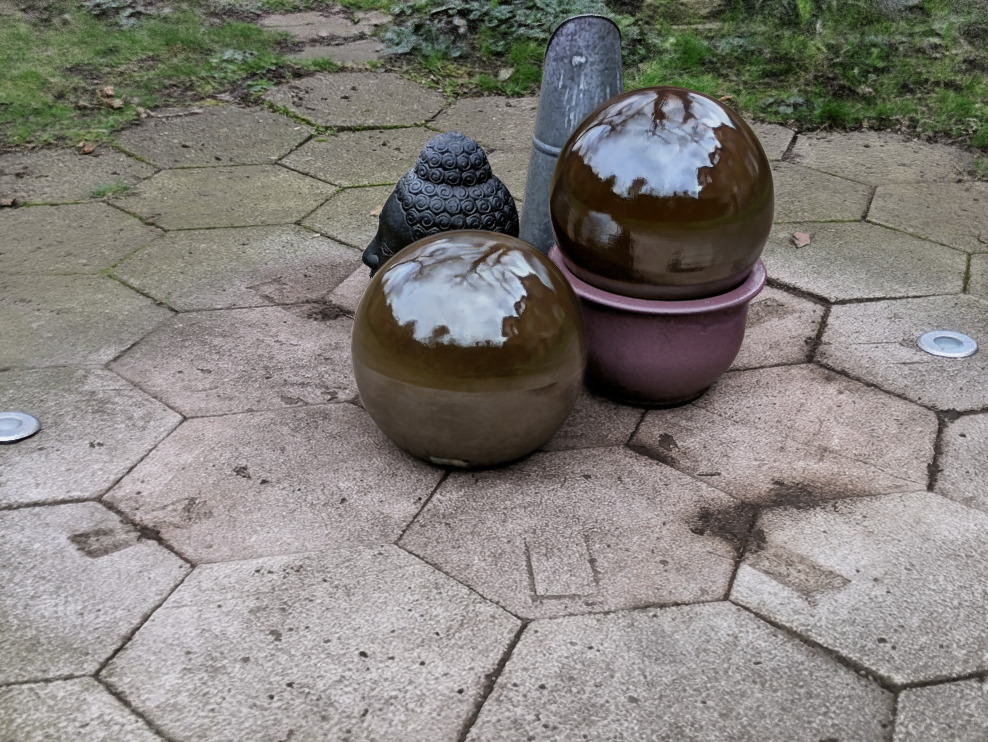} &
        \includegraphics[width=\spherecompwidth, trim={125mm 120mm 75mm 30mm}, clip]{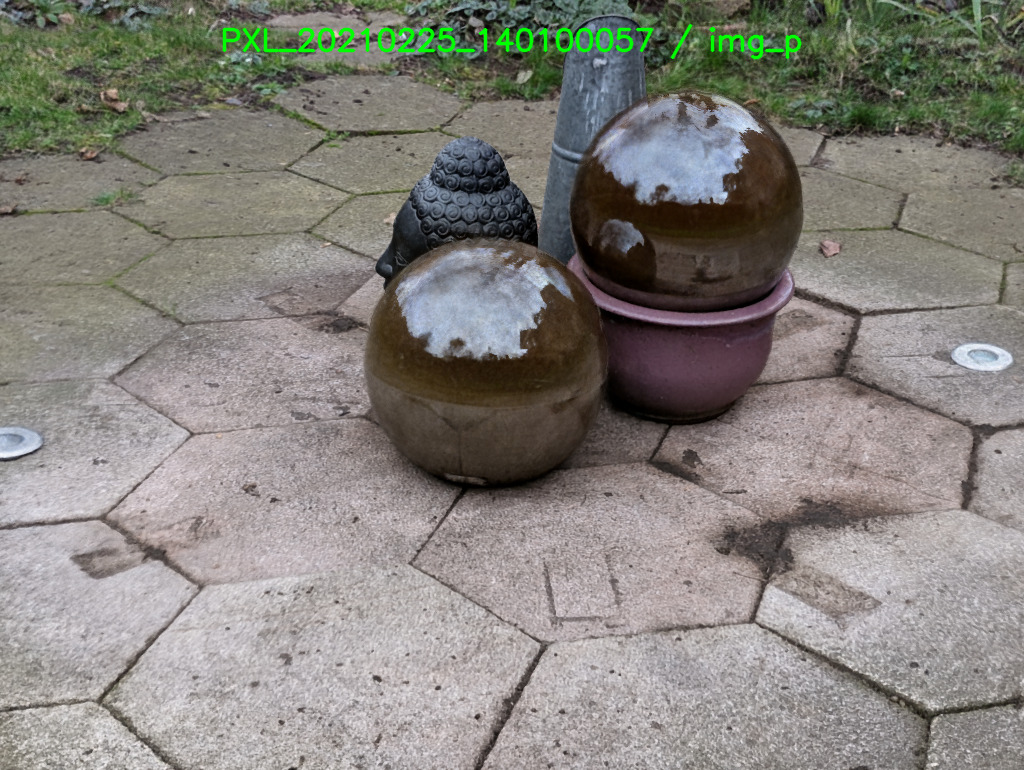} &
        \includegraphics[width=\spherecompwidth, trim={125mm 120mm 75mm 30mm}, clip]{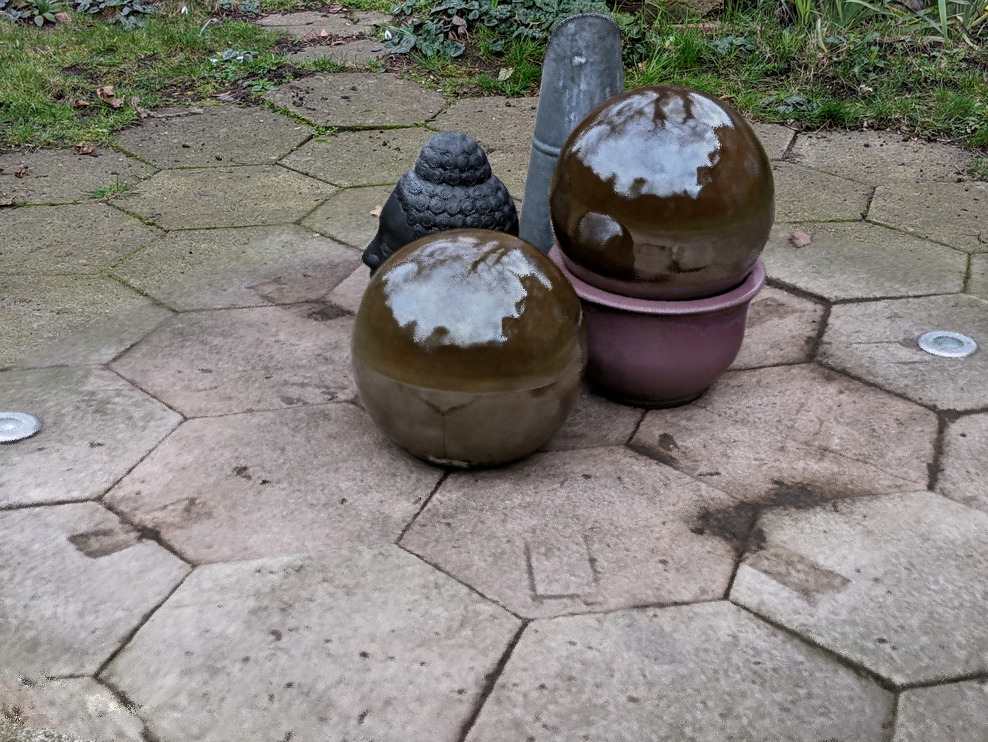} &
        \includegraphics[width=\spherecompwidth, trim={125mm 120mm 75mm 30mm}, clip]{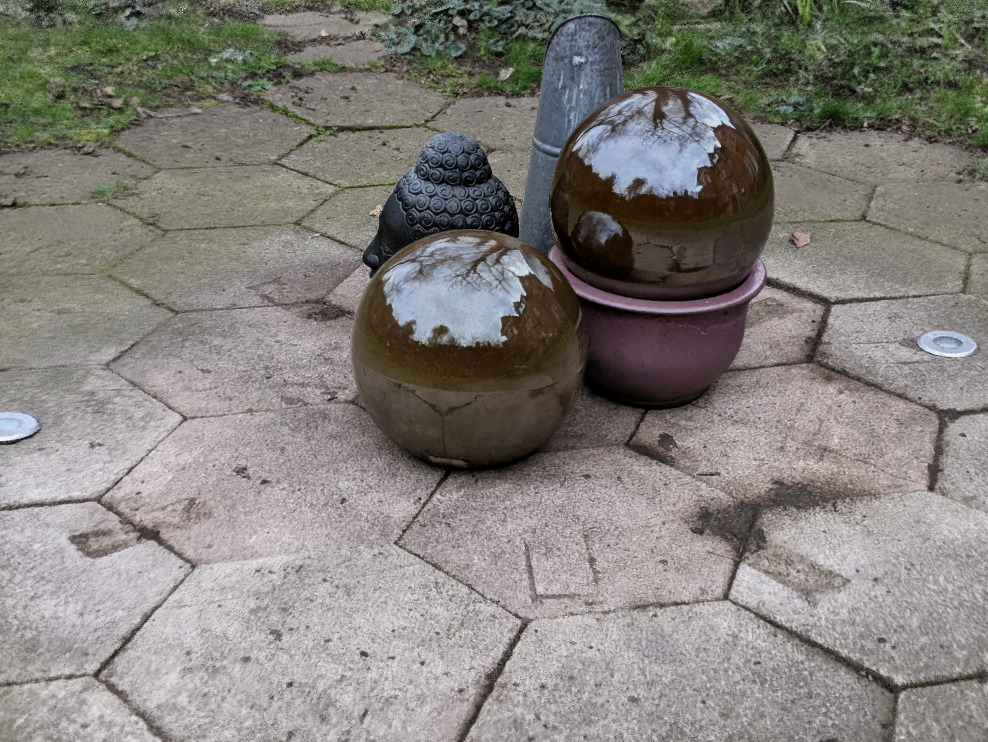} &
        \includegraphics[width=\spherecompwidth, trim={125mm 120mm 75mm 30mm}, clip]{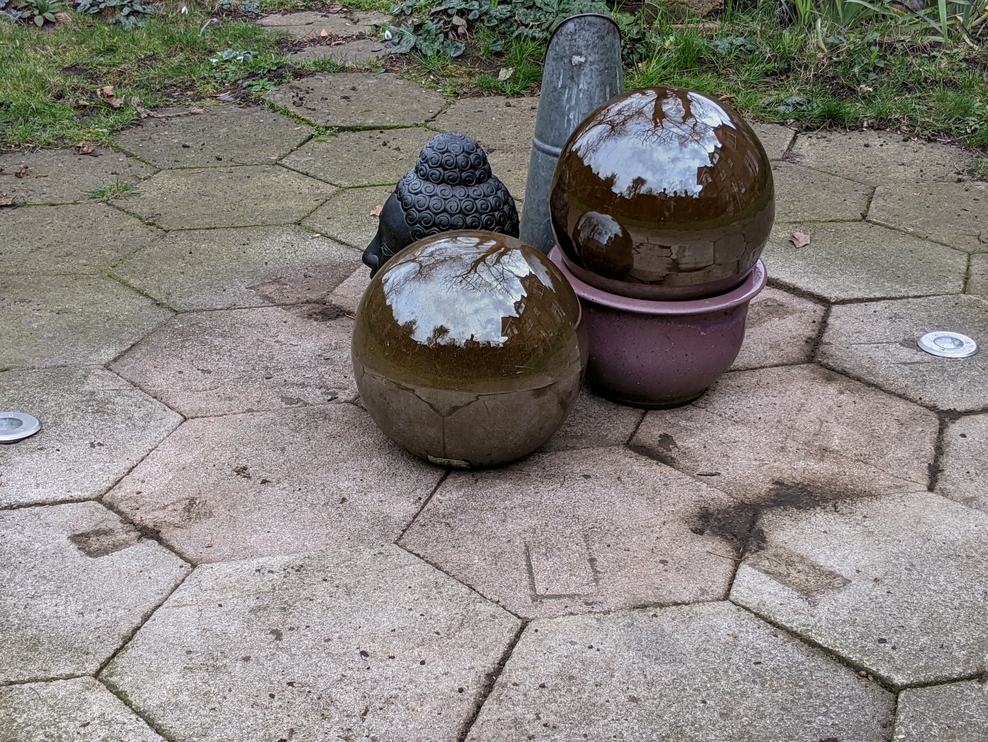}
        \\ 
        \small (a) Zip-NeRF & \small (b) UniSDF & \small (c) SpecNeRF & \small (d) NDE & \small (e) Ours & \small Ground truth
    \end{tabular}
    \vspace{-0.13in}
\caption{
By reflecting rays into the geometry of the model, our method (e) is able to render specular reflections that are more accurate than those rendered by existing techniques (a-d), which tend to be limited by the capacity of their representation of view-dependent appearance. Notice the high-frequency reflections rendered by our model, including sharp details in the tree branches and building reflected in the sphere on the right of the image.
}
\label{fig:spheres_comp}
\end{figure*}

\begin{figure*}
\centering
    \begin{tabular}{@{}c@{\,\,}c@{\,\,}c@{\,\,}c@{}}
        \includegraphics[width=\nearfieldwidth, trim={320px 170px 230px 90px}, clip]{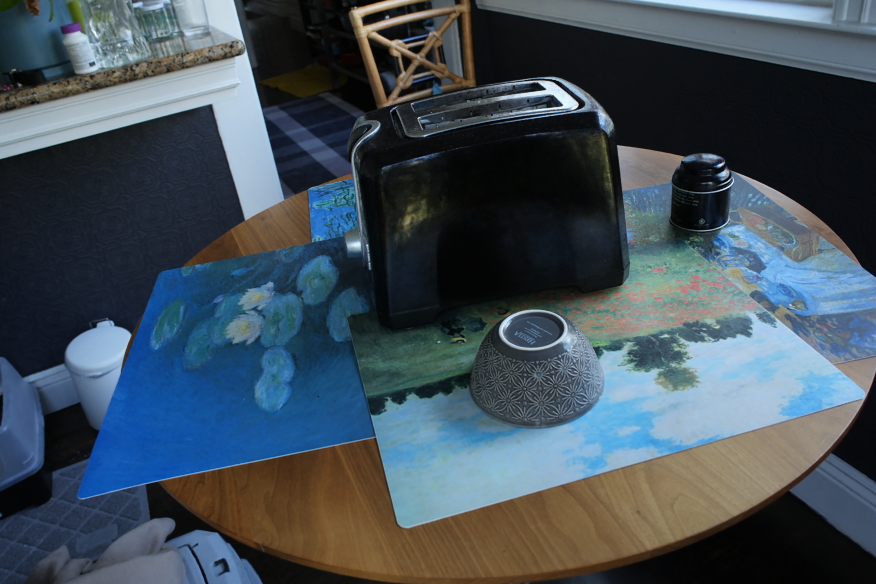} &
        \includegraphics[width=\nearfieldwidth, trim={320px 170px 230px 90px}, clip]{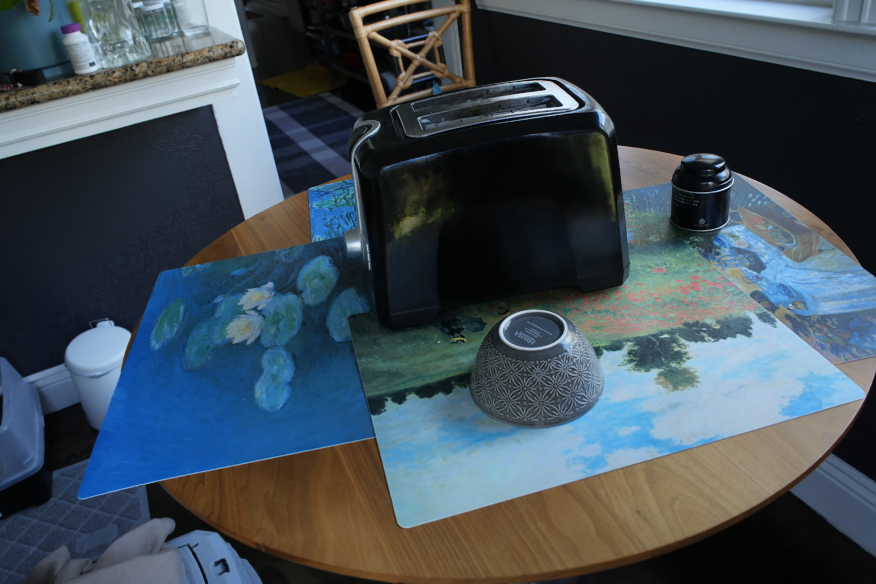} &
        \includegraphics[width=\nearfieldwidth, trim={320px 170px 230px 90px}, clip]{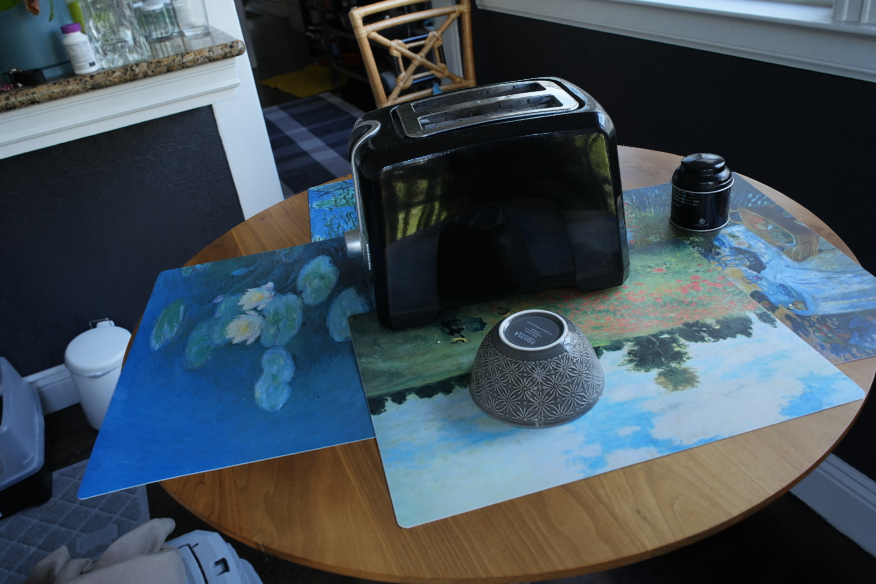} &
        \includegraphics[width=\nearfieldwidth, trim={320px 170px 230px 90px}, clip]{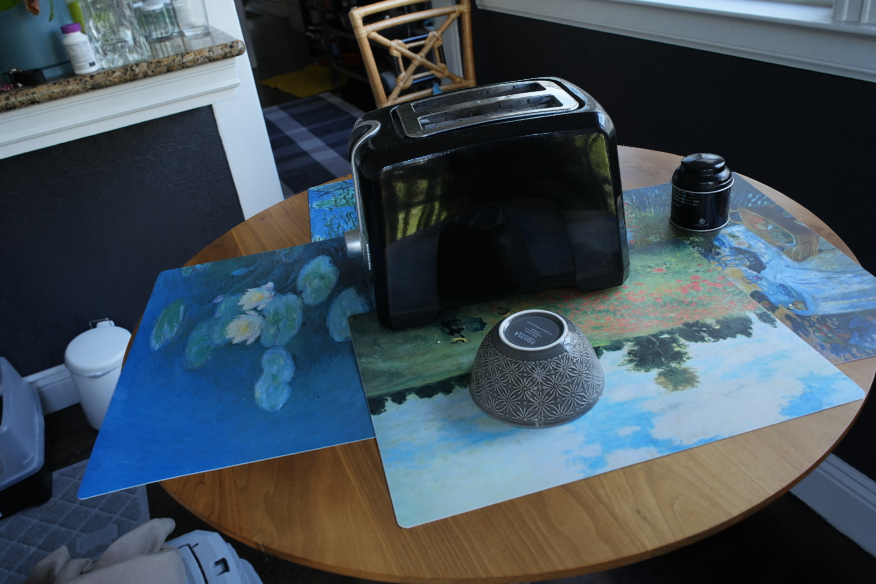} 
        \\ 
        \includegraphics[width=\nearfieldwidth, trim={360px 210px 210px 110px}, clip]{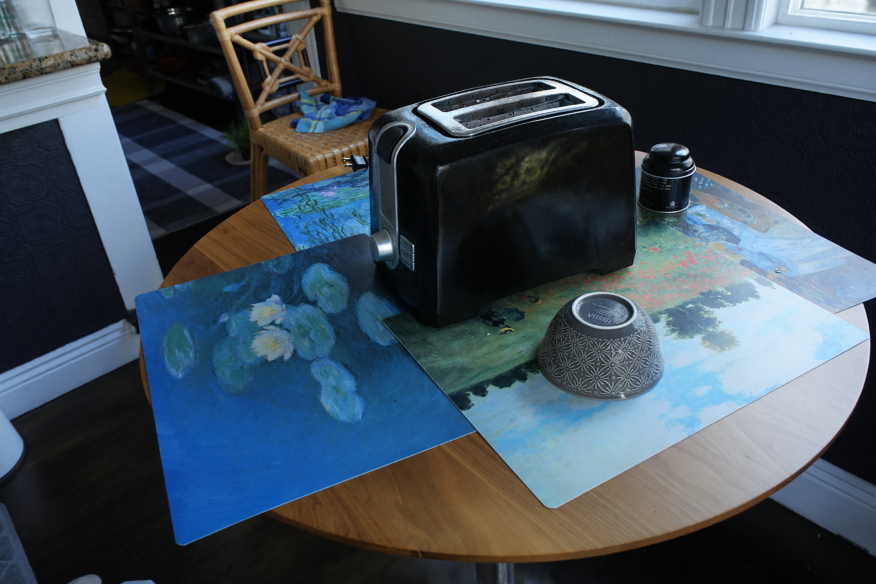} &
        \includegraphics[width=\nearfieldwidth, trim={360px 210px 210px 110px}, clip]{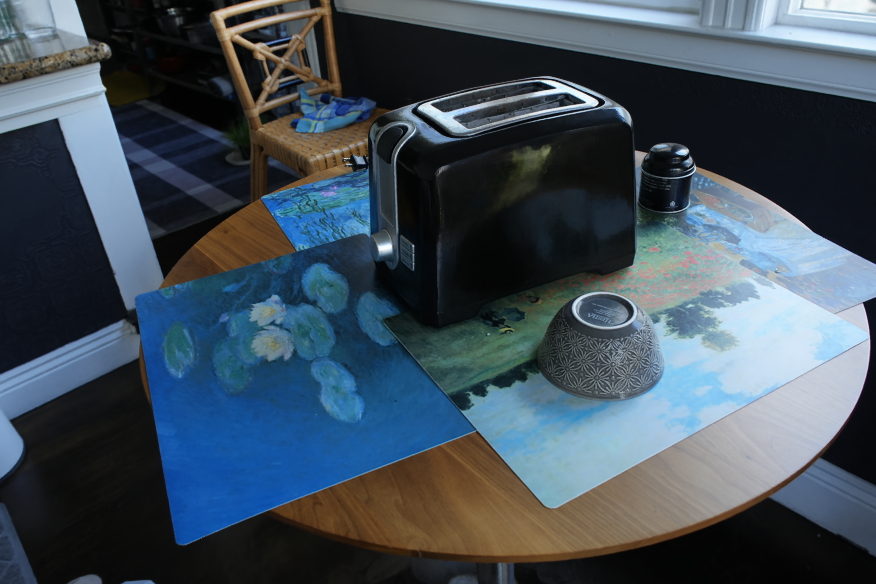} &
        \includegraphics[width=\nearfieldwidth, trim={360px 210px 210px 110px}, clip]{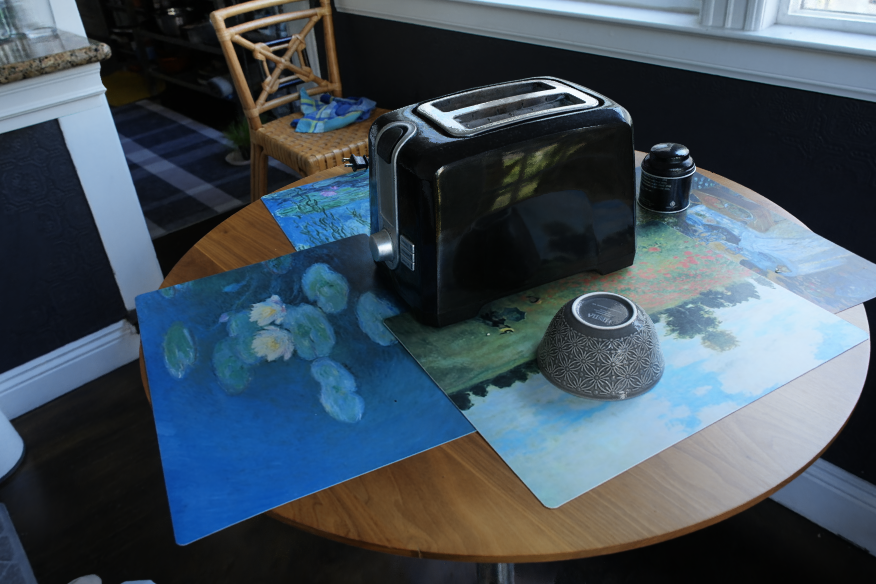} &
        \includegraphics[width=\nearfieldwidth, trim={360px 210px 210px 110px}, clip]{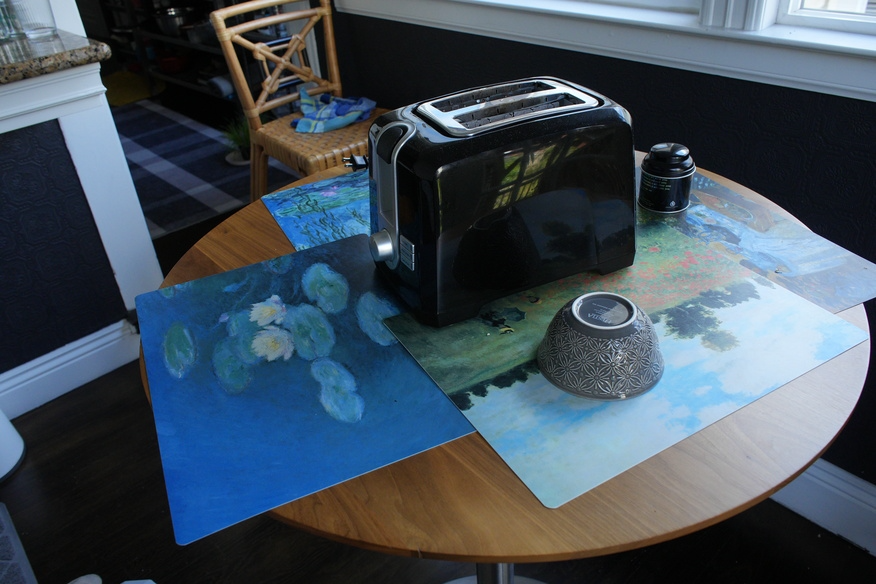} 
        \\
        \small (a) Zip-NeRF & \small (b) UniSDF & \small (c) Ours & \small Ground truth
    \end{tabular}
\vspace{-0.15in}
\caption{
Our model represents detailed and accurate reflections of nearby content. Other baselines that do not explicitly trace reflection rays are unable to render reflections of the window, bowl, and painting in this example. Furthermore, notice how our method accurately models how these near-field reflections move across the toaster as the camera viewpoint changes.
}
\label{fig:toaster_comp}
\end{figure*}

\cleardoublepage

{\huge Supplementary material: \\
NeRF-Casting: Improved View-Dependent Appearance with Consistent Reflections}
\\

A major benefit of our method is the visual quality of results and reflections as the camera moves around the scene. This is best visualized in the set of comparisons included in our interactive supplemental webpage at \MYhref{https://nerf-casting.github.io}{\texttt{https://nerf-casting.github.io}}.

\appendix
\section{Reflected Cone Derivation}
\label{app:reflected_ray}

In Section~\ref{sec:generalenc} we describe our approach of shifting the ray's origin away from the surface point in order to match the reflected ray's  radius with the incident radius. Denote the incident camera ray as $\vr(t) = \vo + t\viewdir$, where its origin and direction are $\vo$ and $\viewdir$ respectively, and denote its radius by $\dot{r}$. Similarly, denote the reflected ray from the point $\avgx$ as $\vr'(t) = \vo' + t\refdir$, with origin and direction $\vo'$ and $\refdir$ respectively, and denote its radius by $\dot{r}'$. As described in mip-NeRF~\cite{barron2021mipnerf}, the radius of the primary ray at $\avgx$ is $\dot{r}\|\avgx - \vo\|$, and the radius of the reflected ray is $\dot{r}'\|\avgx - \vo'\|$. In order to equate the two radii at the intersection point we must satisfy:
\begin{equation}
    \dot{r}\|\avgx - \vo\| = \dot{r}'\|\avgx - \vo'\|\,,
\end{equation}
which along with the fact that $\vo'$ must be on the ray $\vr'(t)$, and with Equation~\ref{eq:vmfconcentration} can be solved to yield Equation~\ref{eq:shiftedorigin}.

\section{Directional Sampling}
\label{app:directional_sampling}

As described in~\cite{kurz2016unscented}, we choose $K$ directions on the sphere $\{\refdir_j\}_{j=1}^{K}$ such that the mean $\refdir$ and concentration parameter $\avgkappa$ of the vMF distribution are maintained:
\begin{align} \label{eq:refdirs}
    \refdir_1 &= \refdir\,,\;\; \text{ and for } j \in \{0, \ldots, K-2\}: \\
    \refdir_{j+2} &= \cos\lft(\psi\rgt)\refdir+\sin\lft(\psi\rgt)\lft(\cos\lft(\frac{2j\pi}{K-1}\rgt)\vt_1+\sin\lft(\frac{2j\pi}{K-1}\rgt)\vt_2\rgt) \nonumber\,,
\end{align}
where $(\vt_1, \vt_2, \refdir)$ form an orthonormal basis, and the angle $\psi$ between the mean reflection direction $\refdir$ and the other $K-1$ offset reflections $\{\refdir_{j}\}_{j=2}^{K}$ is chosen such that:
\begin{equation} \label{eq:psi}
    \cos(\psi) = \frac{K\lft(\coth(\bar{\kappa})-\Large\sfrac{1}{\bar{\kappa}}\rgt) - 1}{K-1}.
\end{equation}

Note that there is rotational symmetry in how the tangent vectors $\vt_1$ and $\vt_2$ are defined. During optimization we randomly rotate them in the tangent plane, and during evaluation we set a deterministic orientation. See the supplement for specific details.

In Section~\ref{sec:aa} we define an orthonormal basis $(\vt_1, \vt_2, \refdir)$ around the reflection direction $\refdir$ defined in Equation~\ref{eq:meanrefray}. The two tangent vectors $\vt_1$ and $\vt_2$ are symmetric and may be rotated arbitrarily about the normal vector, and we do that during optimization. During evaluation, however, we use a deterministic basis by first defining:
\begin{equation} \label{eq:upvec}
    \mathbf{u} = \begin{cases}\hat{\mathbf{z}} \quad \text{if } |\refdir\cdot\hat{\mathbf{z}}|<0.9 \\ \hat{\mathbf{y}} \quad \text{otherwise}\end{cases}
\end{equation}
where $(\hat{\mathbf{x}}, \hat{\mathbf{y}}, \hat{\mathbf{z}})$ are the standard basis for $\mathbb{R}^3$. We then set the tangent vectors to:
\begin{equation}
    \vt_1 = \frac{\mathbf{u}\times\refdir}{\|\mathbf{u}\times\refdir\|}\,,\quad\quad \vt_2 = \refdir\times\vt_1\,.
\end{equation}

This approach is numerically stable for any reflection direction $\refdir$, since the denominator when normalizing $\vt_1$ is bounded from below.

\section{Optimization and Representation Specifications}

In this section we specify our method's architecture and parameterizations, as well as the optimization procedure used to fit it to a collection of images.

\subsection{Parameterization} \label{sec:geometrysupp}

We use the three-round proposal sampling procedure used by mip-NeRF 360 and Zip-NeRF~\cite{barron2022mipnerf360,barron2023zipnerf}. Our $j$th sampling round uses a hash grid with resolution $2^5, ..., 2^{M_i}$, with $M_1 = 9$, $M_2 = 11$, and $M_3 = 13$, each having a maximum number of $2^21$ elements. For camera rays, the first two rounds of sampling use $64$ samples each, and the third uses $32$ samples. For reflected rays, we only use the first two rounds, and generate $64$ samples in each one. Our appearance model uses a hash grid with resolutions $2^0, ..., 2^9$ with two dimensions each.

We find that using separate feature grids for the predicted normals and for roughness is beneficial in our experiments. Therefore, the final round of sampling queries three separate NGP features with eight total dimensions: one for roughness, three for predicted normals, and four for the density, bottleneck vector, and mixing coefficients $\beta$. The roughness values are obtained by applying to the raw NGP feature $\tilde{\vf}_\roughness$ a two-layer MLP with hidden dimension $64$, followed by a softplus activation. We also apply a bias of $-1$ to the features:
\begin{equation}
    \roughness = \operatorname{softplus}\lft(\operatorname{MLP}(\tilde{\vf}_\roughness)-1\rgt)\,.
\end{equation}

The predicted normals are obtained from the raw NGP features $\tilde{\vf}_n$ by applying another two-layer MLP with hidden dimension $64$, the output of which is normalized:
\begin{equation}
    \tilde{\normal} = \operatorname{normalize}\lft(\operatorname{MLP}\lft(\tilde{\vf}_n\rgt)\rgt),
\end{equation}
where $\operatorname{normalize}(\vx) = \nicefrac{\vx}{\|\vx\|}$.

Finally, the remaining features $\tilde{\vf}_b$ are used to output density using another two-layer MLP, followed by an exponential nonlinearity (with a bias of $2$):
\begin{equation}
    \tau = \exp\lft(\operatorname{MLP}\lft(\tilde{\vf}_b\rgt)+2\rgt)\,,
\end{equation}
and to output the mixing coefficients:
\begin{equation}
    \beta = \operatorname{sigmoid}\lft(\operatorname{MLP}\lft(\tilde{\vf}_b\rgt)\rgt)\,,
\end{equation}

The view-dependent color MLP $g$ from Equation~\ref{eq:rgbv} takes as inputs the position $\vx$, bottleneck vector $\vb$, normal $\normal$, and view direction $\viewdir$. The reflection-dependent color MLP $h$ from Equation~\ref{eq:rgbr} takes as input the position $\vx$, bottleneck $\vb$, normal $\normal$, the dot product between the normal and view direction $\normal\cdot\viewdir$, the reflected ray direction $\refdir$, and the feature vector $\avgf$. Additionally, we feed into both $f$ and $g$ the camera position with a low-degree positional encoding of $2$, \ie, we input $\lft(\cos(\vo), \sin(\vo), \cos(2\vo), \sin(\vo)\rgt)$. This is designed to enable the MLP to account for spatial variations in appearance due to transient effects like shadowing and illumination changes---something that can be encoded into view-dependent appearance by models relying on a large-capacity MLP like Zip-NeRF. Both of our MLPs $g$ and $h$ are two-layer MLPs with hidden dimension $128$, followed by a logistic sigmoid nonlinearity.

We apply a coarse-to-fine schedule to all NGP grids corresponding to $\tilde{\vf}_n$, $\tilde{\vf}_b$, and $\tilde{f}_b$, similar to UniSDF~\cite{wang2023unisdf}. Unlike UniSDF, we use an approach more similar to Nerfies~\cite{park2021nerfies}, which uses a cosine-shaped windowing function. Every feature $\vf$ gets multiplied by a cosine window:
\begin{equation}
    \vf' = \vf\odot \mathbf{w}(\boldsymbol{\nu}, t)\,,
\end{equation}
where $\odot$ denotes elementwise multiplication, $\boldsymbol{\nu}$ is a vector of resolutions corresponding to the levels of the hash grid (as also used in the main paper), $t \in [0, 1]$ is a scalar which grows linearly from $0$  to $1$ during optimization, and the windowing function is defined as:
\begin{align}
    \mathbf{w}(\boldsymbol{\nu}, t) &= \frac{1-\cos(\pi \boldsymbol{\xi}(\boldsymbol{\nu}, t))}{2},\\ \text{where }\boldsymbol{\xi}(\boldsymbol{\nu}, t) &= \operatorname{clip}\lft(\log_2(m)-\log_2(\boldsymbol{\nu})-1+st\rgt)\,,
\end{align}
where $s$ and $m$ are hyperparameters controling the rate of introduction of new scales and the initial scale, respectively, and $\operatorname{clip}$ is a function that clips its input to $[0, 1]$. In our experiments we set $m=16\sqrt{2}$ and $s=25$.

\subsection{Optimization}

We inherit Zip-NeRF's optimization schedule. However, due to the increased peak memory consumption of having both primary and reflected rays, we halve our batch size from $2^{16}$ to $2^{15}$, and run optimization for $50$k iterations instead of $25k$. Like Zip-NeRF, we use the Adam optimizer~\cite{adam} with $\beta_1=0.9$, $\beta_2=0.99$, and $\varepsilon=10^{-15}$, and our learning rate is decayed logarithmically from $10^{-2}$ to $10^{-3}$ for the first $25$k iterations after a $5$k cosine warmup.  Unlike Zip-NeRF we continue training for an additional $25$k iterations with a constant learning rate of $10^{-3}$. Because we double the number of iterations while halving the batch size, our model ``sees`` roughly as many rays/pixels as Zip-NeRF during training.

Zip-NeRF performed proposal supervision using a spline-based loss that encouraged proposal distributions along each ray to resemble the final NeRF distribution along the ray. This loss requires a hyperparameter $r$ for each proposal level, which is the radius of a rectangular pulse that the NeRF distribution is blurred by when computing this spline-based loss. Correctly setting this hyperparameter is critical to avoid aliasing along rays, and in Zip-NeRF the radii are manually set to $0.03$ and $0.003$ for the two proposal levels respectively. In our model we use the proposal network not just for rendering pixels using ray intervals of constant length, but also for rendering reflections using ray intervals of \emph{variable} length, so using manually-tuned hyperparameters for these blur radii is challenging. We therefore modify Zip-NeRF's proposal supervision procedure to avoid the need for a manually-set $r$ value, and instead blur each interval of the NeRF distribution by a rectangular pulse whose radius is the weighted geometric mean of the intervals of the proposal distribution that overlap with each NeRF interval. This not only avoids the need to manually specify the $r$ hyperparameter individually for each primary and reflected proposal level, but also generally reduces aliasing by allowing $r$ to be determined adaptively during training and independently for each interval (rather than a single value being used across all intervals). Computing this adaptive radius is straightforward: we compute the logarithm of the width of each interval in the proposal distribution, then use Zip-NeRF's resampling functionality to interpolate into a step function whose values are those log-widths, then exponentiate the average log-width for each NeRF interval to yield a geometric mean.

Like in Zip-NeRF, we regularize the density field using the distortion loss from mip-NeRF 360~\cite{barron2022mipnerf360}. Critically, we apply the distortion loss to both camera rays \emph{and} reflected rays. This prevents the gradients of the reflected ray densities from creating ``floater'' artifacts in regions that are unobserved or observed by a small number of cameras.

We also use the orientation loss from Ref-NeRF~\cite{verbin2022refnerf}, applied directly to the normal vectors corresponding to the density field, with weight $10^{-3}$.

Finally, we apply stop-gradient operators to prevent gradients from affecting the weights when accumulating the normal vectors and intersection points according to Equation~\ref{eq:volrenderxn} of the main paper.

\newcommand{\datawidth}{0.245\linewidth}
\newcommand{\beemerwidth}{0.2832\linewidth}
\begin{figure*}
    \centering
    \begin{tabular}{@{}c@{\,\,}c@{\,\,}c@{\,\,}c@{}}
        \includegraphics[width=\datawidth]{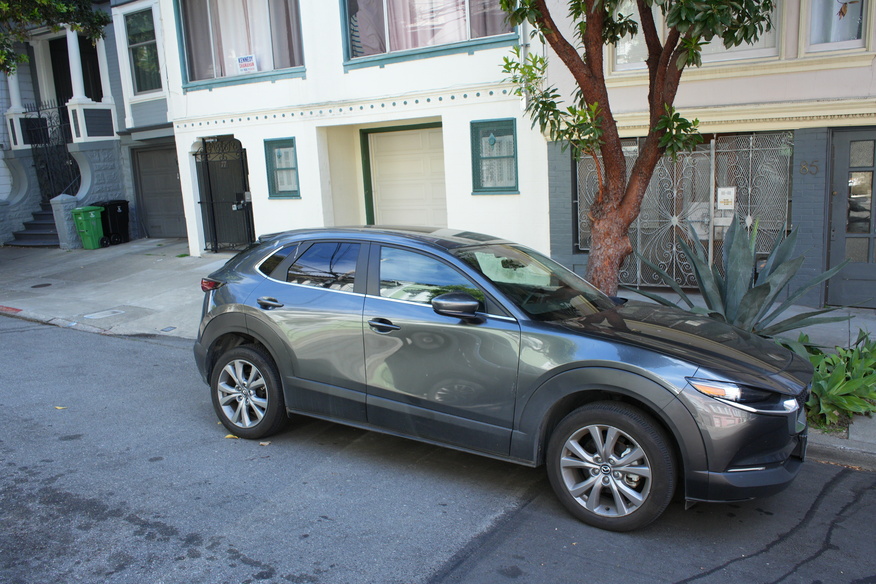} &
        \includegraphics[width=\datawidth]{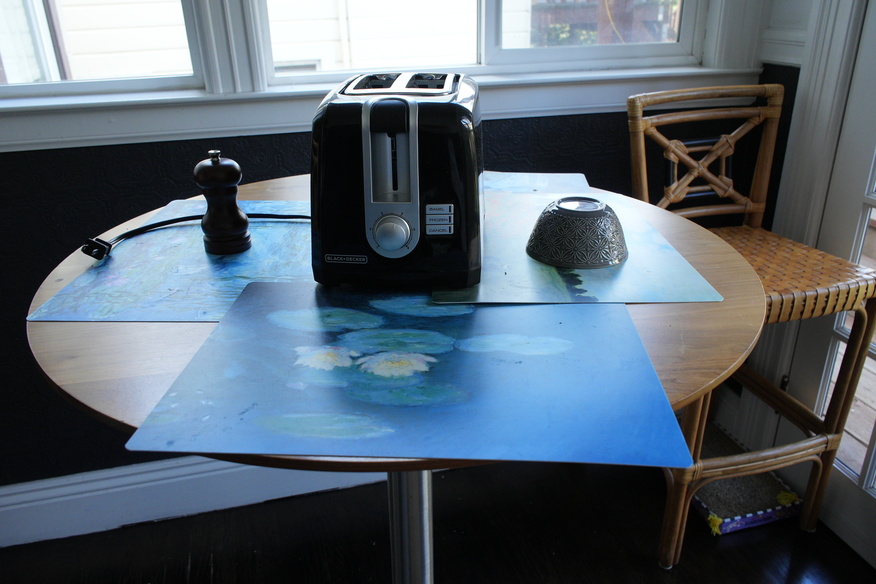} &
        \includegraphics[width=\datawidth]{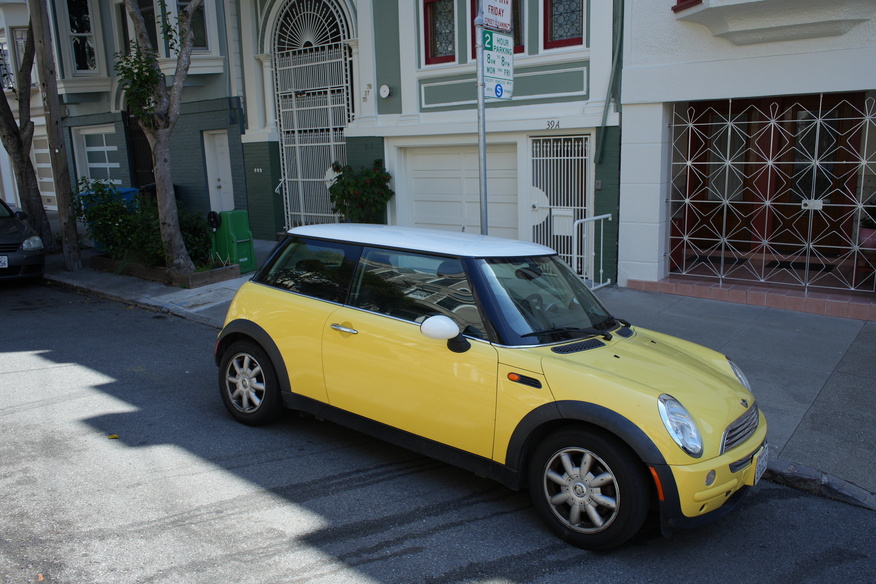} &
        \includegraphics[width=\datawidth]{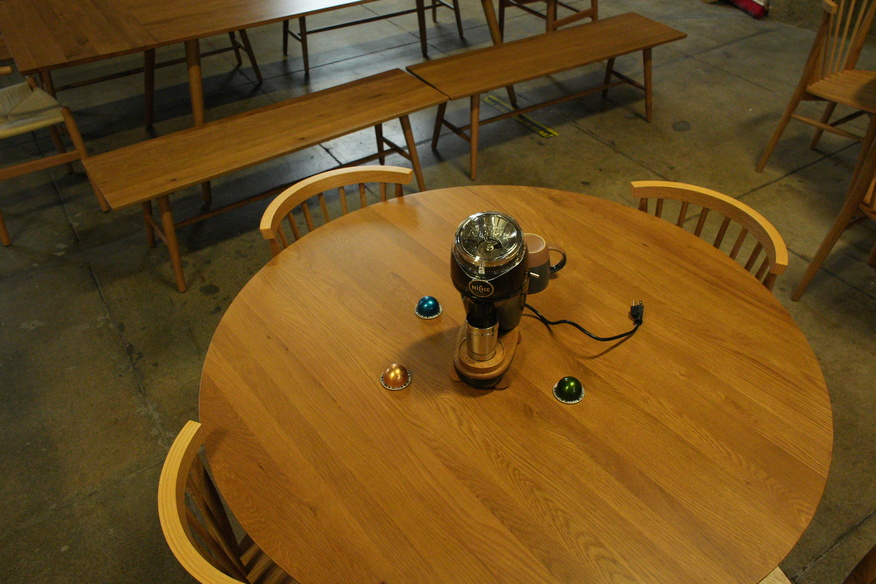}
        \\ 
        \includegraphics[width=\datawidth]{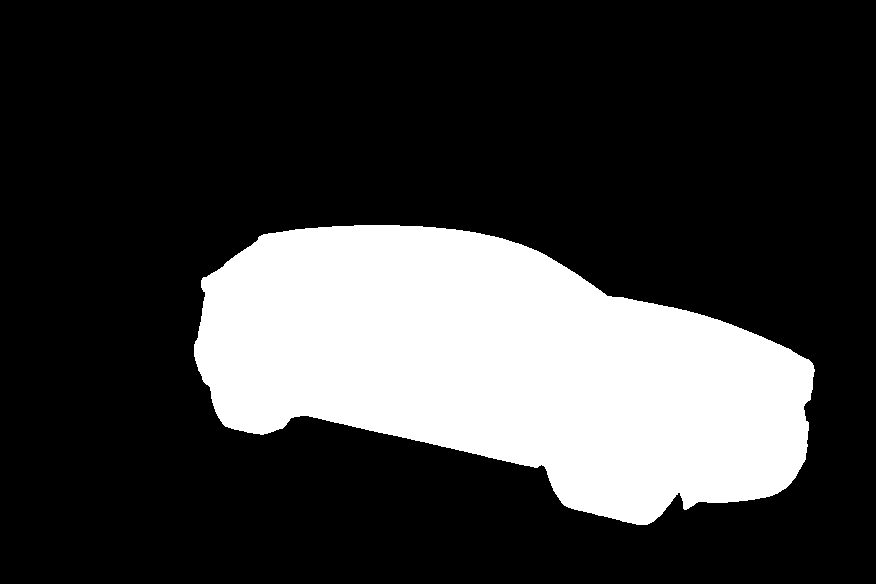} &
        \includegraphics[width=\datawidth]{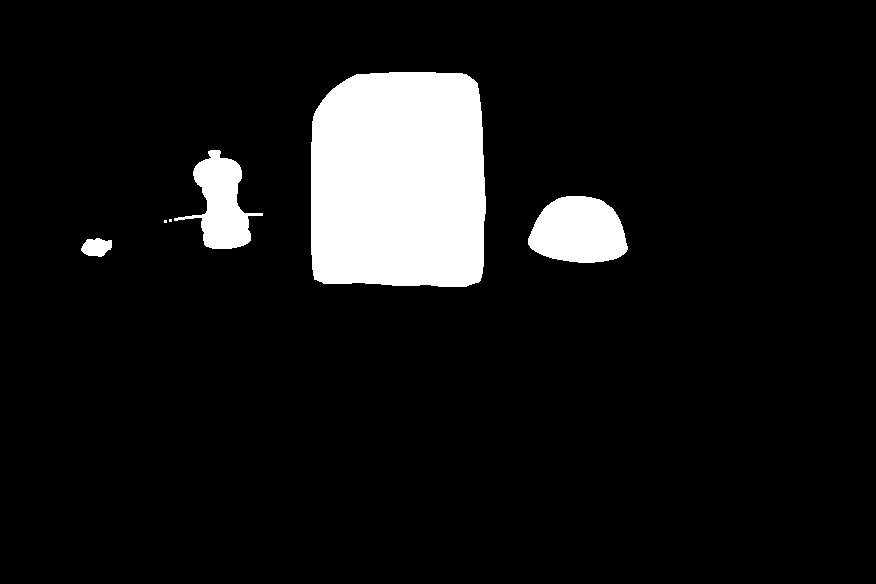} &
        \includegraphics[width=\datawidth]{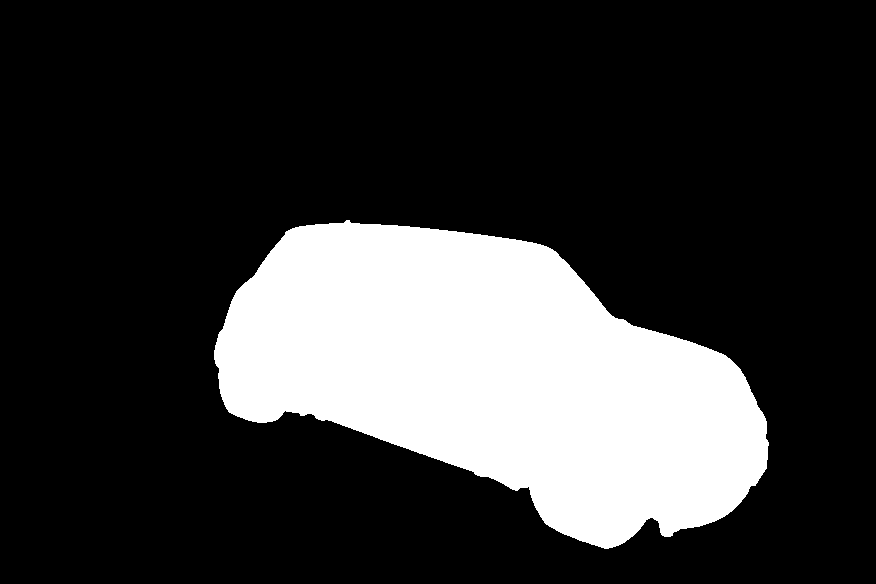} &
        \includegraphics[width=\datawidth]{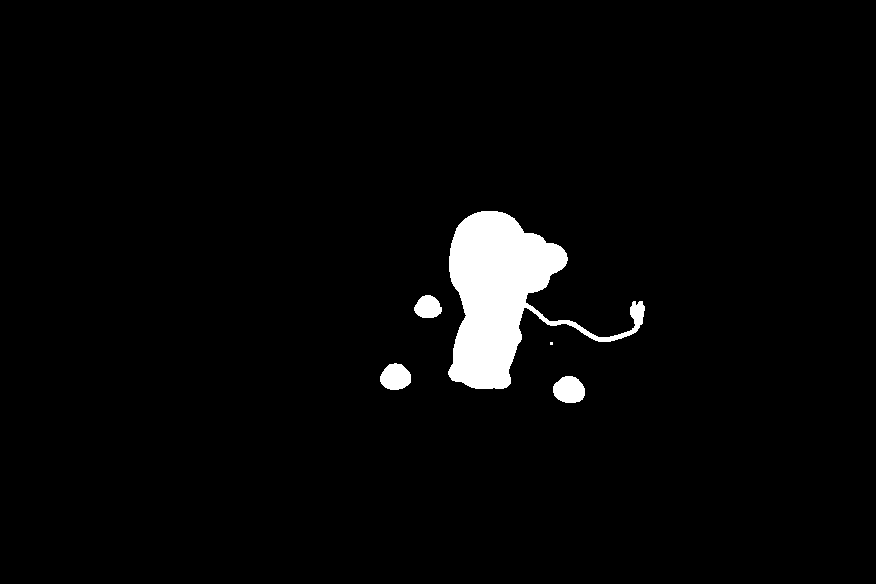}
    \end{tabular}
    \begin{tabular}{@{}c@{\,\,}c@{\,\,}c@{}}
        \includegraphics[width=\datawidth]{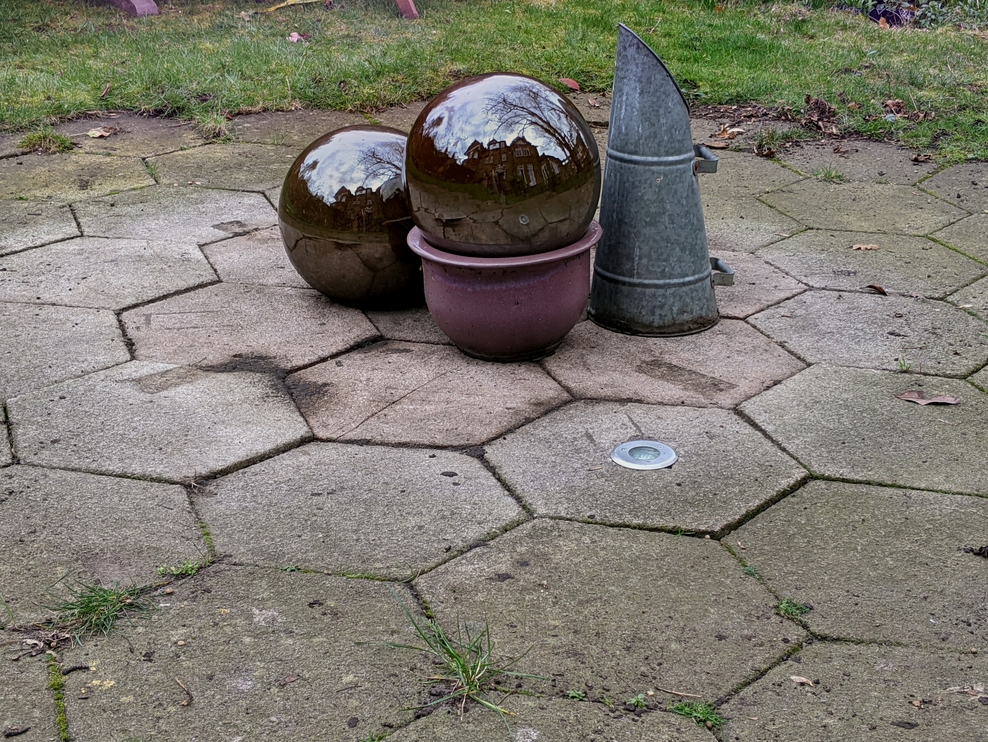} &
        \includegraphics[width=\datawidth]{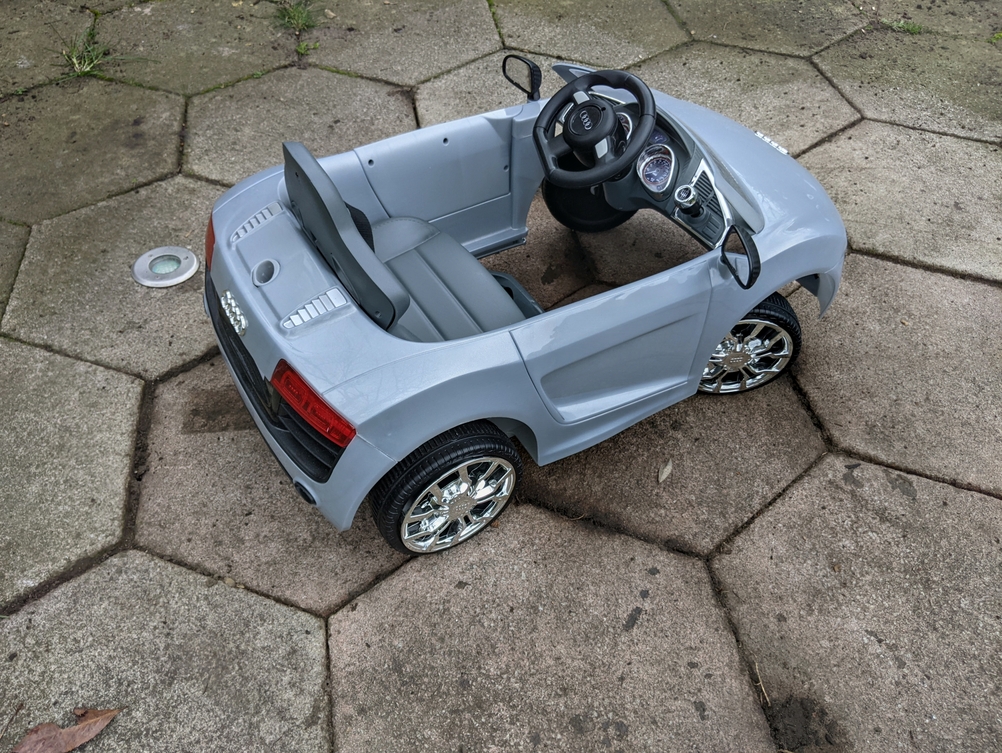} &
        \includegraphics[width=\beemerwidth]{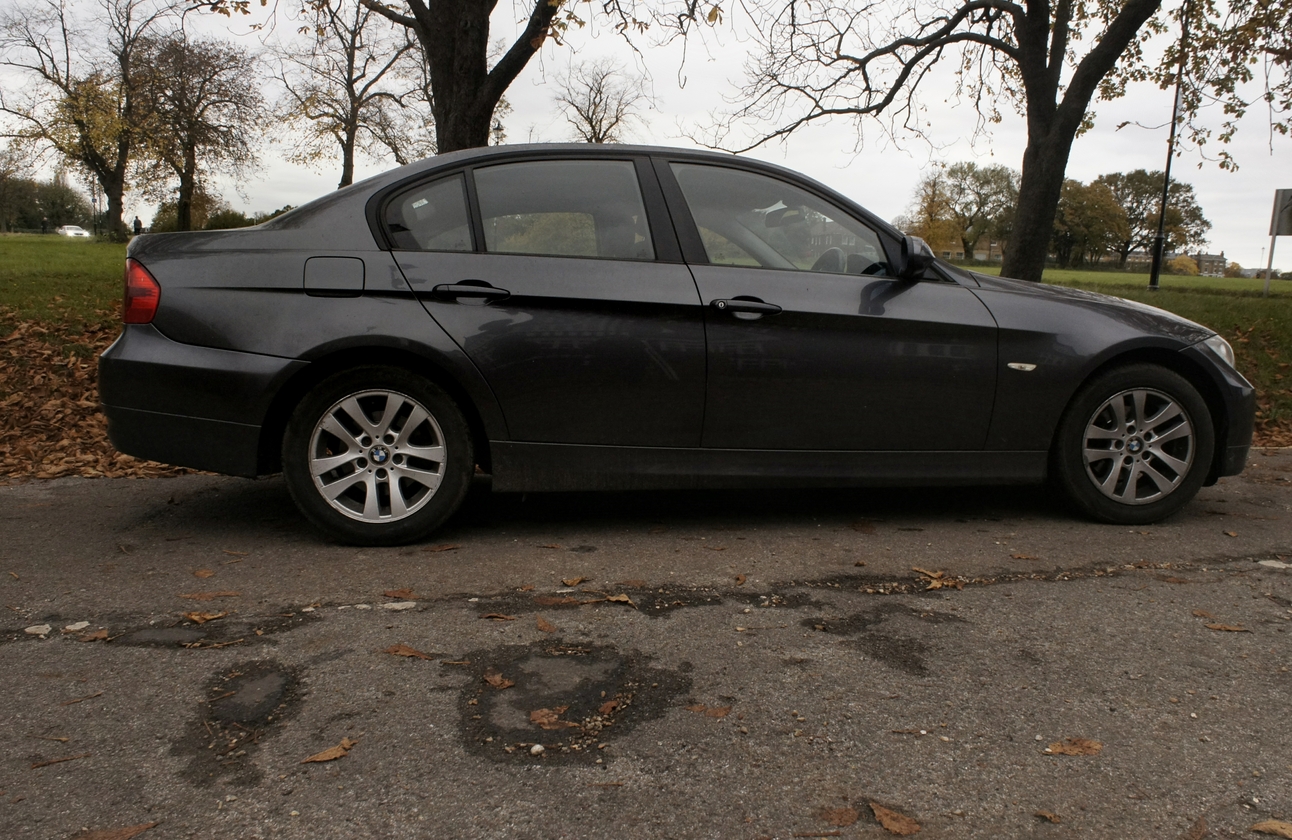}
        \\ 
        \includegraphics[width=\datawidth]{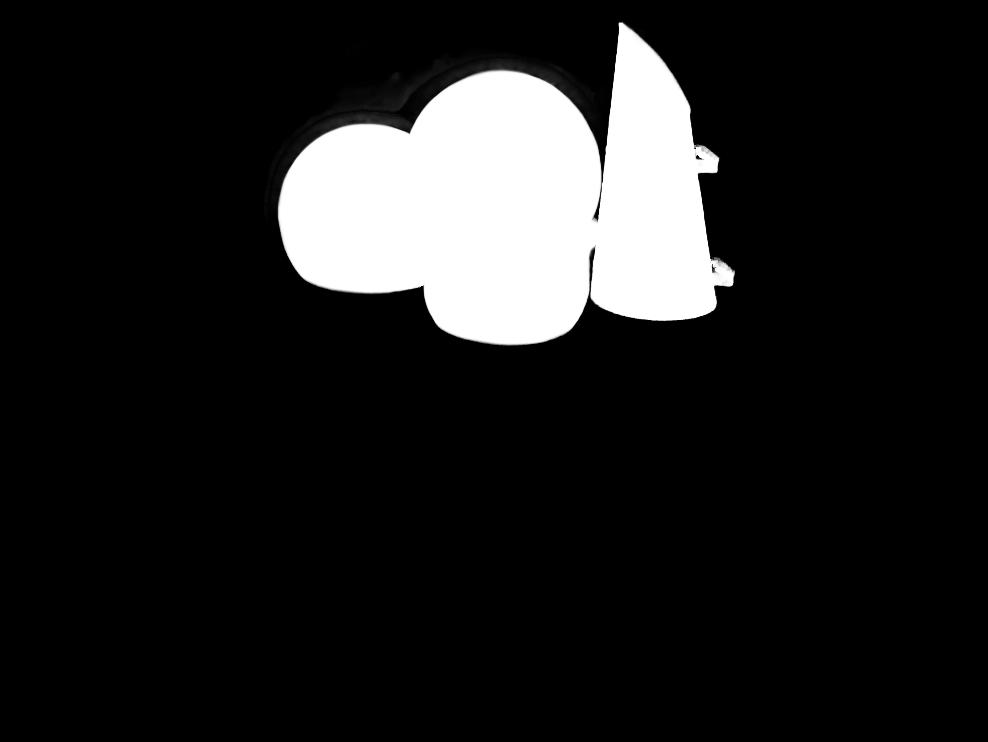} &
        \includegraphics[width=\datawidth]{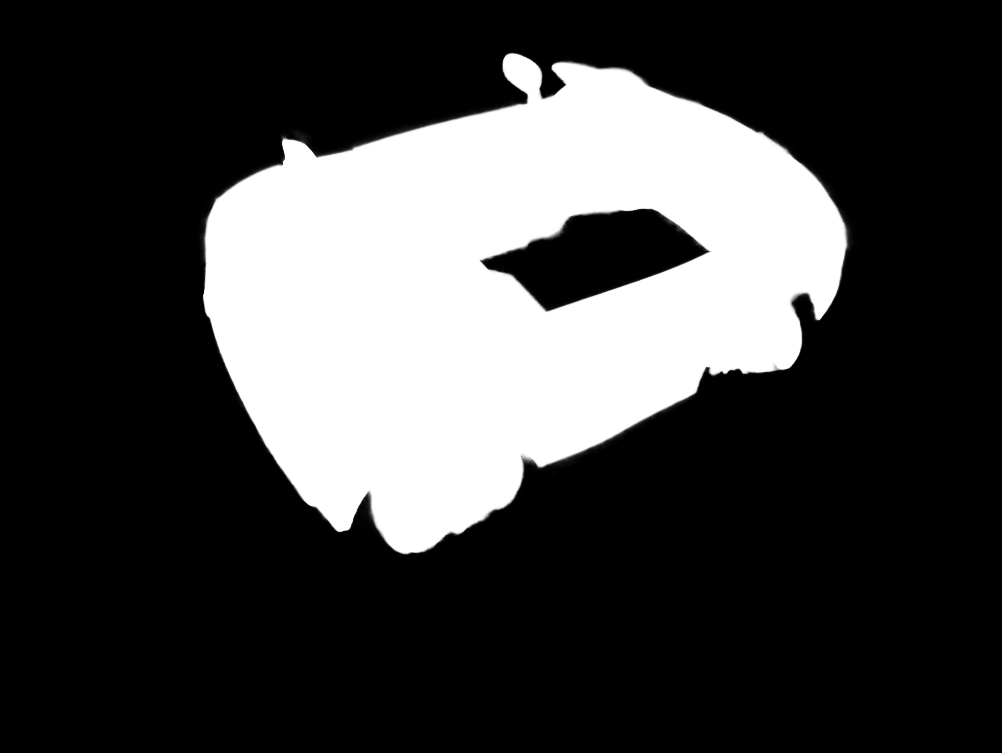} &
        \includegraphics[width=\beemerwidth]{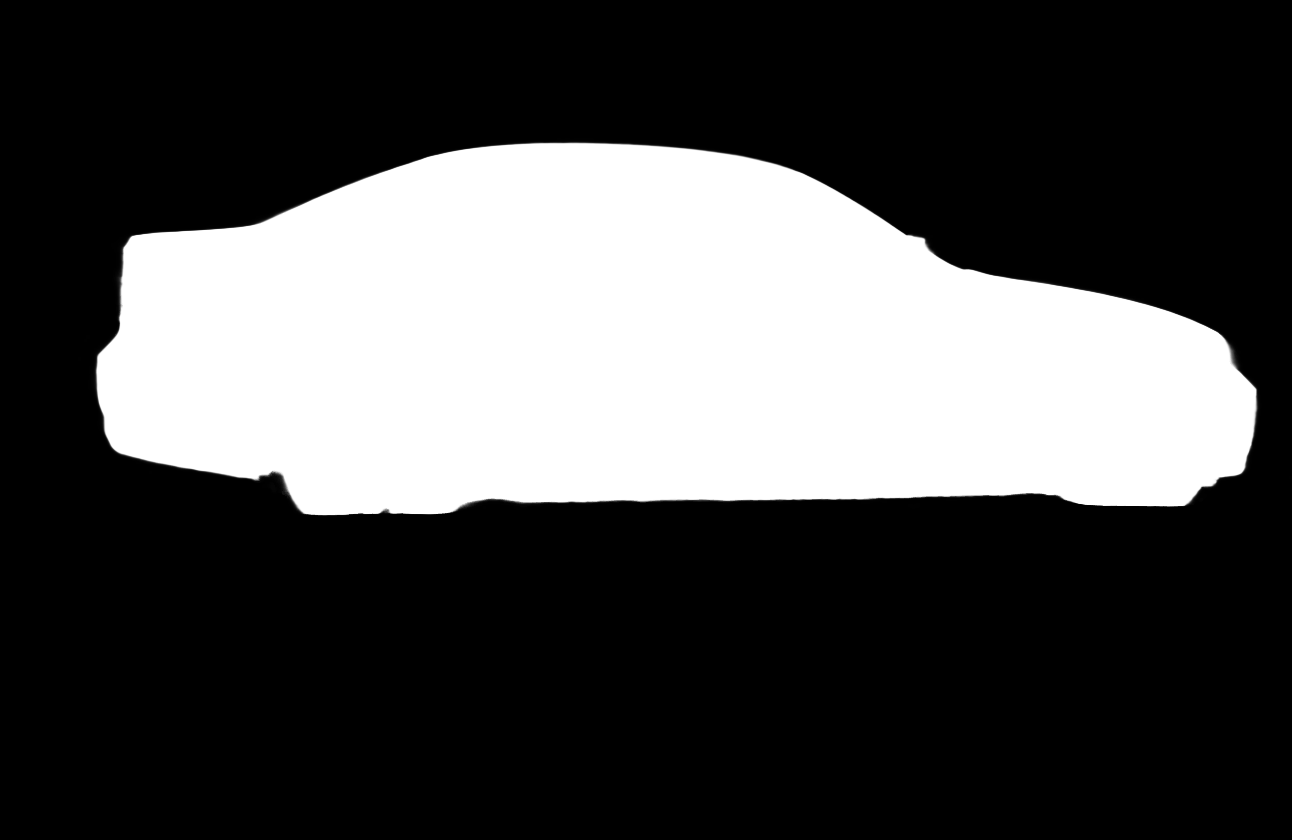}
    \end{tabular}
    \vspace{-0.15in}
    \caption{Images showing our new four scenes (top) along with their per-image masks, and the masks obtained for the real Ref-NeRF data~\cite{verbin2022refnerf}.
    \label{fig:dataset}
    }
\end{figure*}

\subsection{Dataset and Masks} \label{sec:suppdata}

Our newly-captured dataset contains four new scenes with between $231$ and $348$ images each, all with resolution $3504\times 2336$ (which we downsample by a factor of $4$, as done in the mip-NeRF 360 outdoor dataset). Like the mip-NeRF 360 and real Ref-NeRF datasets, we use every 8th frame as a test images, and use the rest for training. The per-scene breakdown is as follows:
\begin{enumerate}
    \item \emph{toaster} contains $348$ images, out of which $44$ are used for testing.
    \item \emph{hatchback} contains $308$ images, $38$ used for testing.
    \item \emph{grinder} contains $231$ images, $28$ used for testing.
    \item \emph{compact} contains $332$ images, $41$ used for testing.
\end{enumerate}

See Figure~\ref{fig:dataset} for examples of images captured from these four scenes.

Additionally, in order to focus on the appearance of reflections in these scenes as well as the captured scenes from the Ref-NeRF dataset, we calculate masked metrics which only take into account foreground pixels belonging to shiny objects. The masks are computed by defining a 3D bounding box for the foreground reflective objects, rendering the opacity within the bounding box for each test view, binarizing the resulting accumulated opacity, and filling small holes. Figure~\ref{fig:dataset} also shows examples of images and their corresponding masks from all seven scenes.

\section{Per-scene Results}

Table~\ref{tab:ablationmasked} presents our per-scene results for Table~\ref{tab:mainresults} of the main paper. Additionally, Table~\ref{tab:ablation} presents the same results for the full, unmasked images.

\begin{table*}[b!]
    \centering
    \begin{tabular}{|l||ccc|cccc|}
    \hline
    \bf{Masked PSNR} & \multicolumn{3}{c|}{Ref-NeRF Real Scenes} & \multicolumn{4}{c|}{Our Shiny Scenes}\\
    \input{tables/ablations_per_masked_psnr}  \!\!\! \\
    \hline
    \multicolumn{7}{c}{} \\
    \hline
    \bf{Masked SSIM} & \multicolumn{3}{c|}{Ref-NeRF Real Scenes} & \multicolumn{4}{c|}{Our Shiny Scenes}\\
    \input{tables/ablations_per_masked_ssim}  \!\!\! \\
    \hline
    \multicolumn{7}{c}{} \\
    \hline
    \bf{Masked LPIPS} & \multicolumn{3}{c|}{Ref-NeRF Real Scenes} & \multicolumn{4}{c|}{Our Shiny Scenes}\\
    \input{tables/ablations_per_masked_lpips}  \!\!\! \\
    \hline
    \end{tabular}
    \caption{Ablation study of our method as well as comparison to prior work, using masked images. The metrics are computed only taking into account pixels belonging to the foreground objects. See Section~\ref{sec:suppdata} for more information and examples. 
    }
    \label{tab:ablationmasked}
\end{table*}

\begin{table*}[b!]
    \centering
    \begin{tabular}{|l||ccc|cccc|}
    \hline
    \bf{PSNR} & \multicolumn{3}{c|}{Ref-NeRF Real Scenes} & \multicolumn{4}{c|}{Our Shiny Scenes}\\
    \input{tables/ablations_per_psnr}  \!\!\! \\
    \hline
    \multicolumn{7}{c}{} \\
    \hline
    \bf{SSIM} & \multicolumn{3}{c|}{Ref-NeRF Real Scenes} & \multicolumn{4}{c|}{Our Shiny Scenes}\\
    \input{tables/ablations_per_ssim}  \!\!\! \\
    \hline
    \multicolumn{7}{c}{} \\
    \hline
    \bf{LPIPS} & \multicolumn{3}{c|}{Ref-NeRF Real Scenes} & \multicolumn{4}{c|}{Our Shiny Scenes}\\
    \input{tables/ablations_per_lpips}  \!\!\! \\
    \hline
    \end{tabular}
    \caption{Ablation study of our method as well as comparison to prior work, using unmasked images.}
    \label{tab:ablation}
\end{table*}

\begin{table*}[b!]
    \centering
    \begin{tabular}{|l||ccccc|cccc|ccc|}
    \hline
    \bf{PSNR} & \multicolumn{5}{c|}{Mip-NeRF 360 Outdoor Scenes} & \multicolumn{4}{c|}{Mip-NeRF 360 Indoor Scenes} & \multicolumn{3}{c|}{Ref-NeRF Real Scenes}\\
    \input{tables/360_per_psnr}  \!\!\! \\
    \hline
    \multicolumn{10}{c}{} \\
    \hline
    \bf{SSIM} & \multicolumn{5}{c|}{Mip-NeRF 360 Outdoor Scenes} & \multicolumn{4}{c|}{Mip-NeRF 360 Indoor Scenes}& \multicolumn{3}{c|}{Ref-NeRF Real Scenes} \\
    \input{tables/360_per_ssim}  \!\!\! \\
    \hline
    \multicolumn{10}{c}{} \\
    \hline
    \bf{LPIPS} & \multicolumn{5}{c|}{Mip-NeRF 360 Outdoor Scenes} & \multicolumn{4}{c|}{Mip-NeRF 360 Indoor Scenes}& \multicolumn{3}{c|}{Ref-NeRF Real Scenes} \\
    \input{tables/360_per_lpips}  \!\!\! \\
    \hline
    \end{tabular}
    \caption{A comparison on the real captures from mip-NeRF 360~\cite{barron2022mipnerf360} and the three shiny real captures from Ref-NeRF~\cite{verbin2022refnerf}. The ``Ref-NeRF*'' baseline is an improved version of Ref-NeRF that uses Zip-NeRF's geometry model, sampling and optimization procedure, but with Ref-NeRF's appearance model.
    }
    \label{tab:scene_360_results}
\end{table*}

\section{Synthetic Results}

Table~\ref{tab:synthetic_results} shows synthetic results on the \emph{materials} and \emph{hotdog} scenes from the Blender dataset~\cite{mildenhall2020nerf}, which contain strong specularities, and the six shiny scenes from Ref-NeRF~\cite{verbin2022refnerf}. The results demonstrate that despite being designed for robustness on real data, our approach still obtains state-of-the-art results on these shiny synthetic scenes.

\begin{table*}[b!]
    \centering
    \resizebox{0.7\linewidth}{!}{
    \begin{tabular}{|l||cc||cccccc|c|}
    \hline
    \bf{PSNR} & \multicolumn{2}{c||}{Blender Scenes} & \multicolumn{7}{c|}{Shiny Blender Scenes}\\
    \input{tables/synthetic_per_psnr}  \!\!\! \\
    \hline
    \hline
    \bf{SSIM} & \multicolumn{2}{c||}{Blender Scenes} & \multicolumn{7}{c|}{Shiny Blender Scenes}\\
    \input{tables/synthetic_per_ssim}  \!\!\! \\
    \hline
    \hline
    \bf{LPIPS} & \multicolumn{2}{c||}{Blender Scenes} & \multicolumn{7}{c|}{Shiny Blender Scenes}\\
    \input{tables/synthetic_per_lpips}  \!\!\! \\
    \hline
    \end{tabular}
    }
    \caption{
    The per-scene and average metrics over the two scenes from the \emph{Blender} dataset~\cite{mildenhall2020nerf} which contain strong specular effects, and the six shiny scenes from the \emph{Shiny Blender} dataset~\cite{verbin2022refnerf}. The average is taken only over the six \emph{Shiny Blender} scenes.
    }
    \label{tab:synthetic_results}
\end{table*}

%% file: tables/ablations_per_masked_psnr.tex
 & \textit{sedan} & \textit{toycar} & \textit{spheres} & \textit{hatchback} & \textit{toaster} & \textit{compact} & \textit{grinder} \\\hline
UniSDF                      &                   31.58 &                   28.09 & \cellcolor{orange}32.33 &                   30.98 & \cellcolor{orange}37.66 &                   33.95 & \cellcolor{orange}39.07 \\
Zip-NeRF                    &                   30.83 &                   28.04 &                   29.38 &                   31.30 &                   36.85 &    \cellcolor{red}35.04 &    \cellcolor{red}40.19 \\
Ref-NeRF*                    & 30.74 & 28.13 & 29.08 & 29.51 & 37.02 & 34.42 & \cellcolor{yellow}39.20 \\
Ours with single downweighted cone  & \cellcolor{yellow}33.01 & \cellcolor{yellow}28.87 &                   31.68 &                   31.37 &                   37.41 &                   34.15 &                   38.89 \\
Ours with single dilated cone       &                   32.59 & \cellcolor{orange}28.88 &                   31.93 &    \cellcolor{red}32.22 &                   36.90 & \cellcolor{orange}34.58 &                   38.91 \\
Ours without downweighting          &                   32.77 & \cellcolor{yellow}28.87 &                   31.79 &                   31.57 &                   37.54 &                   33.97 &                   38.93 \\
Ours with 3D Jacobian               & \cellcolor{orange}33.06 &                   28.77 &                   30.86 & \cellcolor{yellow}31.81 & \cellcolor{yellow}37.61 & \cellcolor{yellow}34.55 &                   38.87 \\
Ours without near-field reflections &                   31.89 &    \cellcolor{red}28.91 & \cellcolor{yellow}32.32 &                   31.19 &                   36.69 &                   33.41 &                   38.70 \\
Ours                                &    \cellcolor{red}33.30 & \cellcolor{yellow}28.87 &    \cellcolor{red}32.80 & \cellcolor{orange}31.97 &    \cellcolor{red}37.87 &                   33.55 & 38.99

%% file: tables/ablations_per_masked_ssim.tex
 & \textit{sedan} & \textit{toycar} & \textit{spheres} & \textit{hatchback} & \textit{toaster} & \textit{compact} & \textit{grinder} \\\hline
UniSDF                      &                   0.941 &                   0.930 & \cellcolor{yellow}0.955 &                   0.952 & \cellcolor{yellow}0.986 &                   0.968 &    \cellcolor{red}0.994 \\
Zip-NeRF                    &                   0.936 & \cellcolor{yellow}0.934 &                   0.944 &                   0.953 &                   0.985 &    \cellcolor{red}0.972 &    \cellcolor{red}0.994 \\
Ref-NeRF*                    & 0.936 & 0.932 & 0.941 & 0.943 & 0.983 & \cellcolor{yellow} 0.970 & \cellcolor{red}0.994 \\
Ours with single downweighted cone  & \cellcolor{orange}0.949 &    \cellcolor{red}0.937 &                   0.952 &                   0.955 & \cellcolor{yellow}0.986 &    \cellcolor{red}0.972 &    \cellcolor{red}0.994 \\
Ours with single dilated cone       &                   0.947 &    \cellcolor{red}0.937 & \cellcolor{orange}0.956 &    \cellcolor{red}0.961 &                   0.984 &    \cellcolor{red}0.972 &    \cellcolor{red}0.994 \\
Ours without downweighting          & \cellcolor{yellow}0.948 &    \cellcolor{red}0.937 &                   0.954 &                   0.956 & \cellcolor{yellow}0.986 & \cellcolor{orange}0.970 &    \cellcolor{red}0.994 \\
Ours with 3D Jacobian               & \cellcolor{orange}0.949 & \cellcolor{orange}0.936 &                   0.946 & \cellcolor{yellow}0.957 & \cellcolor{orange}0.987 &    \cellcolor{red}0.972 &    \cellcolor{red}0.994 \\
Ours without near-field reflections &                   0.943 &    \cellcolor{red}0.937 & \cellcolor{orange}0.956 &                   0.952 &                   0.984 &                   0.967 & \cellcolor{orange}0.993 \\
Ours                                &    \cellcolor{red}0.950 & \cellcolor{orange}0.936 &    \cellcolor{red}0.962 & \cellcolor{orange}0.959 &    \cellcolor{red}0.988 & 0.969 &    \cellcolor{red}0.994

%% file: tables/ablations_per_masked_lpips.tex
 & \textit{sedan} & \textit{toycar} & \textit{spheres} & \textit{hatchback} & \textit{toaster} & \textit{compact} & \textit{grinder} \\\hline
UniSDF                      &                   0.083 &                   0.064 &                   0.040 &                   0.046 &                   0.022 &                   0.033 & \cellcolor{orange}0.007 \\
Zip-NeRF                    &                   0.082 & \cellcolor{yellow}0.062 &                   0.045 &                   0.045 &                   0.022 &    \cellcolor{red}0.027 &    \cellcolor{red}0.006 \\
Ref-NeRF*                    & 0.082 & 0.066 & 0.046 & 0.052 & 0.024 & \cellcolor{orange}0.029 & \cellcolor{red}0.006 \\
Ours with single downweighted cone  & \cellcolor{orange}0.071 &    \cellcolor{red}0.060 &                   0.039 &                   0.044 & \cellcolor{yellow}0.021 &    \cellcolor{red}0.027 &    \cellcolor{red}0.006 \\
Ours with single dilated cone       &                   0.073 & \cellcolor{orange}0.061 & \cellcolor{yellow}0.037 &    \cellcolor{red}0.039 &                   0.023 &    \cellcolor{red}0.027 &    \cellcolor{red}0.006 \\
Ours without downweighting          & \cellcolor{yellow}0.072 &    \cellcolor{red}0.060 & \cellcolor{yellow}0.037 &                   0.044 & \cellcolor{yellow}0.021 & \cellcolor{orange}0.029 &    \cellcolor{red}0.006 \\
Ours with 3D Jacobian               & \cellcolor{orange}0.071 & \cellcolor{orange}0.061 &                   0.045 & \cellcolor{yellow}0.043 & \cellcolor{orange}0.019 &    \cellcolor{red}0.027 &    \cellcolor{red}0.006 \\
Ours without near-field reflections &                   0.075 &    \cellcolor{red}0.060 & \cellcolor{orange}0.036 &                   0.047 &                   0.024 &                   0.032 & \cellcolor{orange}0.007 \\
Ours                                &    \cellcolor{red}0.070 & \cellcolor{orange}0.061 &    \cellcolor{red}0.030 & \cellcolor{orange}0.040 &    \cellcolor{red}0.018 & \cellcolor{yellow}0.030 &    \cellcolor{red}0.006

%% file: tables/ablations_per_psnr.tex
 & \textit{sedan} & \textit{toycar} & \textit{spheres} & \textit{hatchback} & \textit{toaster} & \textit{compact} & \textit{grinder} \\ \hline
UniSDF                      &                   24.68 &                   24.15 &                   22.27 &                   27.01 &                   32.90 &                   29.72 &                   33.72 \\
Zip-NeRF                    &                   25.85 &                   23.41 &                   21.77 &    \cellcolor{red}27.78 &    \cellcolor{red}33.41 &    \cellcolor{red}31.10 &    \cellcolor{red}34.67 \\
Ref-NeRF*                    & 25.39 & 22.75 & 21.12 & 25.21 & 32.66 & \cellcolor{orange}30.55 & 33.91 \\
Ours with single downweighted cone  & \cellcolor{orange}26.73 &                   24.22 &                   22.85 &                   27.35 &                   32.89 & \cellcolor{yellow}30.01 &                   33.94 \\
Ours with single dilated cone       &                   26.25 & \cellcolor{yellow}24.26 & \cellcolor{yellow}22.94 & \cellcolor{orange}27.57 &                   32.51 &                   29.91 & \cellcolor{yellow}33.98 \\
Ours without downweighting          & \cellcolor{yellow}26.59 &    \cellcolor{red}24.28 &                   22.89 &                   27.29 & \cellcolor{orange}33.07 &                   29.87 &                   33.95 \\
Ours with 3D Jacobian               &    \cellcolor{red}26.77 &                   24.18 &                   22.62 &                   27.39 & \cellcolor{yellow}32.92 & 29.98 &                   33.93 \\
Ours without near-field reflections &                   26.28 & \cellcolor{orange}24.27 & \cellcolor{orange}22.97 &                   27.19 &                   32.54 &                   29.54 &                   33.89 \\
Ours                                &    \cellcolor{red}26.77 &                   24.20 &    \cellcolor{red}23.04 & \cellcolor{yellow}27.49 &                   32.87 &                   29.73 & \cellcolor{orange}34.00

%% file: tables/ablations_per_ssim.tex
 & \textit{sedan} & \textit{toycar} & \textit{spheres} & \textit{hatchback} & \textit{toaster} & \textit{compact} & \textit{grinder} \\\hline
UniSDF                      &                   0.700 &                   0.639 &                   0.567 & \cellcolor{yellow}0.845 & \cellcolor{yellow}0.937 & \cellcolor{yellow}0.895 &                   0.879 \\
Zip-NeRF                    &                   0.733 &                   0.626 &                   0.545 &    \cellcolor{red}0.870 &    \cellcolor{red}0.944 &    \cellcolor{red}0.913 &    \cellcolor{red}0.887 \\
Ref-NeRF*                    & 0.721 & 0.612 & 0.542 & 0.842 & 0.932 & \cellcolor{orange}0.907 & 0.880 \\
Ours with single downweighted cone  &    \cellcolor{red}0.741 & \cellcolor{orange}0.643 &                   0.590 &                   0.841 &                   0.936 &  0.889 & \cellcolor{yellow}0.881 \\
Ours with single dilated cone       &                   0.722 &    \cellcolor{red}0.644 & \cellcolor{orange}0.595 &                   0.834 &                   0.934 &                   0.879 & \cellcolor{yellow}0.881 \\
Ours without downweighting          & \cellcolor{yellow}0.735 & \cellcolor{orange}0.643 &                   0.591 &                   0.842 & \cellcolor{yellow}0.937 &                   0.884 & \cellcolor{yellow}0.881 \\
Ours with 3D Jacobian               &    \cellcolor{red}0.741 &                   0.640 &                   0.581 &                   0.834 & \cellcolor{yellow}0.937 &                   0.884 & \cellcolor{yellow}0.881 \\
Ours without near-field reflections &                   0.731 &    \cellcolor{red}0.644 & \cellcolor{yellow}0.593 &                   0.843 &                   0.935 &                   0.877 & \cellcolor{yellow}0.881 \\
Ours                                & \cellcolor{orange}0.739 & \cellcolor{yellow}0.641 &    \cellcolor{red}0.597 & \cellcolor{orange}0.853 & \cellcolor{orange}0.938 &                   0.884 & \cellcolor{orange}0.882

%% file: tables/ablations_per_lpips.tex
 & \textit{sedan} & \textit{toycar} & \textit{balls} & \textit{hatchback} & \textit{toaster} & \textit{compact} & \textit{grinder} \\\hline
UniSDF                      &                   0.309 &                   0.245 &                   0.243 & 0.160 &                   0.107 & \cellcolor{yellow}0.122 &                   0.132 \\
Zip-NeRF                    &                   0.260 &    \cellcolor{red}0.243 & \cellcolor{orange}0.238 &    \cellcolor{red}0.130 &    \cellcolor{red}0.082 &    \cellcolor{red}0.096 &    \cellcolor{red}0.111 \\
Ref-NeRF*                    & 0.270 & 0.257 & 0.257 & \cellcolor{yellow}0.156 & 0.111 & \cellcolor{orange}0.105 & 0.123 \\
Ours with single downweighted cone  & \cellcolor{yellow}0.256 & \cellcolor{orange}0.245 &                   0.242 &                   0.165 &                   0.102 & 0.142 & \cellcolor{orange}0.114 \\
Ours with single dilated cone       &                   0.271 & \cellcolor{yellow}0.246 & \cellcolor{yellow}0.240 &                   0.172 &                   0.102 &                   0.155 & \cellcolor{yellow}0.115 \\
Ours without downweighting          &                   0.260 & \cellcolor{yellow}0.246 &                   0.242 &                   0.166 & \cellcolor{yellow}0.098 &                   0.151 & \cellcolor{orange}0.114 \\
Ours with 3D Jacobian               & \cellcolor{orange}0.255 &                   0.247 &                   0.251 &                   0.179 &                   0.100 &                   0.155 & \cellcolor{yellow}0.115 \\
Ours without near-field reflections &                   0.261 & \cellcolor{yellow}0.246 &                   0.243 &                   0.164 &                   0.101 &                   0.161 & \cellcolor{yellow}0.115 \\
Ours                                &    \cellcolor{red}0.254 & \cellcolor{yellow}0.246 &    \cellcolor{red}0.238 & \cellcolor{orange}0.155 & \cellcolor{orange}0.096 &                   0.148 & \cellcolor{orange}0.114

%% file: tables/360_per_psnr.tex
 & \textit{bicycle} & \textit{flowers} & \textit{garden} & \textit{stump} & \textit{treehill} & \textit{room} & \textit{counter} & \textit{kitchen} & \textit{bonsai} & \textit{sedan} & \textit{toycar} & \textit{spheres} \\\hline
3DGS      & \cellcolor{orange}25.25 &                   21.52 &  27.41 & \cellcolor{orange}26.55 &                   22.49 &                   30.63 &                   28.70 &                   30.32 &                   31.98 &    25.24 & \cellcolor{yellow}23.91 &                   \cellcolor{yellow} 21.95 \\
UniSDF      &                   24.67 & \cellcolor{orange}21.83 & \cellcolor{orange}27.46 & \cellcolor{yellow}26.39 & \cellcolor{orange}23.51 & \cellcolor{yellow}31.25 & \cellcolor{orange}29.26 & \cellcolor{yellow}31.73 & \cellcolor{yellow}32.86 &                   24.68 & \cellcolor{orange}24.15 & \cellcolor{orange}22.27 \\
Zip-NeRF &    \cellcolor{red}25.80 &    \cellcolor{red}22.40 &    \cellcolor{red}28.20 &    \cellcolor{red}27.55 &    \cellcolor{red}23.89 &    \cellcolor{red}32.65 &    \cellcolor{red}29.38 &    \cellcolor{red}32.50 &    \cellcolor{red}34.46 & \cellcolor{orange}25.85 &                   23.41 & 21.77 \\
Ref-NeRF* & 24.91 &                   21.63 & \cellcolor{yellow}27.45 &                   25.91 &                   21.79 & \cellcolor{orange}31.68 &                   26.02 &                   31.61 &                   32.29 & \cellcolor{yellow} 25.40 &                   22.75 &                   21.13 \\
Ours                              & \cellcolor{yellow}24.92 & \cellcolor{yellow}21.75 &                   27.31 &                   25.64 & \cellcolor{yellow}23.22 & 31.66 & \cellcolor{yellow}28.84 & \cellcolor{orange}32.26 & \cellcolor{orange}33.81 &    \cellcolor{red}26.77 &    \cellcolor{red}24.20 &    \cellcolor{red}23.04

%% file: tables/360_per_ssim.tex
 & \textit{bicycle} & \textit{flowers} & \textit{garden} & \textit{stump} & \textit{treehill} & \textit{room} & \textit{counter} & \textit{kitchen} & \textit{bonsai} & \textit{sedan} & \textit{toycar} & \textit{spheres} \\\hline
3DGS      &    \cellcolor{red}0.771 & \cellcolor{yellow}0.605 &    \cellcolor{red}0.868 & \cellcolor{orange}0.775 &                   0.638 & \cellcolor{orange}0.914 &    \cellcolor{red}0.905 & \cellcolor{yellow}0.922 &                   0.938 &    0.713 & \cellcolor{yellow} 0.636 & \cellcolor{orange} 0.573 \\
UniSDF      &                   0.737 & \cellcolor{orange}0.606 &  0.844 & \cellcolor{yellow}0.759 & \cellcolor{orange}0.670 & \cellcolor{orange}0.914 & \cellcolor{yellow}0.888 &                   0.919 & \cellcolor{yellow}0.939 &                   0.700 & \cellcolor{orange}0.639 & \cellcolor{yellow}0.567 \\
Zip-NeRF & \cellcolor{orange}0.769 &    \cellcolor{red}0.642 & \cellcolor{orange}0.860 &    \cellcolor{red}0.800 &    \cellcolor{red}0.681 &    \cellcolor{red}0.925 & \cellcolor{orange}0.902 &    \cellcolor{red}0.928 &    \cellcolor{red}0.949 & \cellcolor{orange}0.733 &                   0.626 &                   0.545 \\
Ref-NeRF* &                   0.723 &                   0.592 & \cellcolor{yellow}0.845 &                   0.731 &                   0.634 & \cellcolor{orange}0.914 &                   0.875 & \cellcolor{yellow}0.922 &                   0.935 &  \cellcolor{yellow} 0.722 &                   0.612 &                   0.542 \\
Ours                              & \cellcolor{yellow}0.747 & \cellcolor{yellow}0.605 &                   0.836 &                   0.749 & \cellcolor{yellow}0.653 & \cellcolor{yellow}0.911 &                   0.887 & \cellcolor{orange}0.924 & \cellcolor{orange}0.945 & \cellcolor{red}0.739 &    \cellcolor{red}0.641 &    \cellcolor{red}0.597

%% file: tables/360_per_lpips.tex
 & \textit{bicycle} & \textit{flowers} & \textit{garden} & \textit{stump} & \textit{treehill} & \textit{room} & \textit{counter} & \textit{kitchen} & \textit{bonsai} & \textit{sedan} & \textit{toycar} & \textit{spheres} \\\hline
3DGS      &    \cellcolor{red}0.205 &                   0.336 &    \cellcolor{red}0.103 & \cellcolor{orange}0.210 &                   0.317 &                   0.220 & \cellcolor{yellow}0.204 &                   0.129 &                   0.205 &   0.301 &  \cellcolor{red} 0.237 & \cellcolor{yellow} 0.248 \\
UniSDF      &                   0.243 & 0.320 &  0.136 & \cellcolor{yellow}0.242 & \cellcolor{orange}0.265 & \cellcolor{orange}0.206 &                   0.206 & 0.124 & 0.184 &                   0.309 & \cellcolor{yellow}0.245 & \cellcolor{orange}0.243 \\
Zip-NeRF & \cellcolor{orange}0.208 &    \cellcolor{red}0.273 & \cellcolor{orange}0.118 &    \cellcolor{red}0.193 &    \cellcolor{red}0.242 &    \cellcolor{red}0.196 &    \cellcolor{red}0.185 &    \cellcolor{red}0.116 &    \cellcolor{red}0.173 & \cellcolor{orange}0.260 & \cellcolor{orange}0.243 &    \cellcolor{red}0.238 \\
Ref-NeRF* &                   0.256 & \cellcolor{yellow} 0.317 & \cellcolor{yellow}0.132 &                   0.261 &                   0.294 & \cellcolor{orange}0.206 &                   0.213 & \cellcolor{yellow}0.121 & \cellcolor{yellow}0.182 &   \cellcolor{yellow} 0.270 &                   0.257 &                   0.257 \\
Ours                              & \cellcolor{yellow}0.231 & \cellcolor{orange}0.312 &                   0.142 &                   0.244 & \cellcolor{yellow}0.273 & \cellcolor{yellow}0.216 & \cellcolor{orange}0.203 & \cellcolor{orange}0.118 & \cellcolor{orange}0.176 & \cellcolor{red}0.254 &                   0.246 &    \cellcolor{red}0.238

%% file: tables/synthetic_per_psnr.tex
 & \textit{hotdog} & \textit{materials} & \textit{teapot} & \textit{toaster} & \textit{car} & \textit{ball} & \textit{coffee} & \textit{helmet} & \textit{mean} \\\hline
Ref-NeRF                & \cellcolor{orange}37.72 &    \cellcolor{red}35.41 & \cellcolor{yellow}47.90 &                   25.70 &    \cellcolor{red}30.82 &    \cellcolor{red}47.46 & \cellcolor{orange}34.21 &                   29.68    & \cellcolor{yellow}35.96\\
ENVIDR                  & --- & \cellcolor{yellow}29.51 &                   46.14 &     \cellcolor{red}26.63 &                   29.88 &                   41.03 &    \cellcolor{red}34.45 & \cellcolor{yellow}36.98 & 35.85\\
Zip-NeRF  & \cellcolor{yellow}37.16 & 31.64 &                   45.66 &                   24.84 &                   27.60 &                   25.30 &                   31.05 &                   27.50 & 30.33 \\
UniSDF                  & --- &                   --- & \cellcolor{orange}48.76 &   \cellcolor{yellow}26.18 &                   29.86 & \cellcolor{yellow}44.10 &                   33.17 & \cellcolor{orange}38.84  & \cellcolor{orange}36.82 \\
Ours &    \cellcolor{red}38.61 & \cellcolor{orange}32.02 &    \cellcolor{red}49.98 & \cellcolor{orange}26.19 & \cellcolor{orange}30.45 & \cellcolor{orange}45.46 & \cellcolor{yellow}33.18 &    \cellcolor{red}39.10 & \cellcolor{red} 37.39

%% file: tables/synthetic_per_ssim.tex
 & \textit{hotdog} & \textit{materials} & \textit{teapot} & \textit{toaster} & \textit{car} & \textit{ball} & \textit{coffee} & \textit{helmet} & \textit{mean} \\\hline
Ref-NeRF                & \cellcolor{orange}0.984 &    \cellcolor{red}0.983 & \cellcolor{orange}0.998 &                   0.922 & \cellcolor{yellow}0.955 & \cellcolor{orange}0.995 & \cellcolor{orange}0.974 &                   0.958 & 0.967 \\
ENVIDR                  & --- & \cellcolor{yellow}0.971 &    \cellcolor{red}0.999 &    \cellcolor{red}0.955 &    \cellcolor{red}0.972 &    \cellcolor{red}0.997 &    \cellcolor{red}0.984 &    \cellcolor{red}0.993 & \cellcolor{red}0.983 \\
Zip-NeRF  &    \cellcolor{orange}0.984 &  0.969 & \cellcolor{yellow}0.997 &                   0.920 &                   0.934 &                   0.928 &                   0.967 &                   0.950 &  0.949\\
UniSDF                  &                     --- &                     --- & \cellcolor{orange}0.998 & \cellcolor{yellow}0.945 &                   0.954 &                   0.993 & \cellcolor{yellow}0.973 & \cellcolor{orange}0.990 & \cellcolor{yellow}0.976 \\
Ours &    \cellcolor{red}0.988 & \cellcolor{orange}0.975 &    \cellcolor{red}0.999 & \cellcolor{orange}0.950 & \cellcolor{orange}0.964 & \cellcolor{yellow}0.994 & \cellcolor{yellow}0.973 & \cellcolor{yellow}0.988 & \cellcolor{orange} 0.977

%% file: tables/synthetic_per_lpips.tex
 & \textit{hotdog} & \textit{materials} & \textit{teapot} & \textit{toaster} & \textit{car} & \textit{ball} & \textit{coffee} & \textit{helmet} & \textit{mean} \\\hline
Ref-NeRF                & \cellcolor{yellow}0.022 &    \cellcolor{red}0.022 & \cellcolor{yellow}0.004 & 0.095 & \cellcolor{yellow}0.041 &                   0.059 & \cellcolor{yellow}0.078 &                   0.075 & 0.059 \\
ENVIDR                  &                   --- & \cellcolor{orange}0.026 & \cellcolor{orange}0.003 &                   0.097 &    \cellcolor{red}0.031 &    \cellcolor{red}0.020 &    \cellcolor{red}0.044 & \cellcolor{yellow}0.022 & \cellcolor{red} 0.036 \\
Zip-NeRF  & \cellcolor{orange}0.020 & 0.032 &  0.007 & \cellcolor{yellow} 0.092 &                   0.055 &                   0.190 & 0.088 &                   0.081 & 0.086 \\
UniSDF                  &                     --- &                   --- & \cellcolor{yellow}0.004 &    \cellcolor{red}0.072 &                   0.047 & \cellcolor{orange}0.039 & \cellcolor{yellow}0.078 & \cellcolor{orange}0.021 & \cellcolor{yellow} 0.044 \\
Ours & \cellcolor{red}0.014 & \cellcolor{yellow}0.027 &    \cellcolor{red}0.002 & \cellcolor{orange}0.073 & \cellcolor{orange}0.033 & \cellcolor{yellow}0.044 & \cellcolor{orange}0.074 &    \cellcolor{red}0.018& \cellcolor{orange} 0.041